\documentclass[runningheads]{llncs}
\usepackage{amsmath}
\usepackage{amssymb}
\usepackage{booktabs} 
\usepackage{graphicx}
\usepackage{pgfplots}
\usepackage[all]{nowidow}
\usepackage[utf8]{inputenc}
\usepackage{multicol}
\usepackage{algpseudocode,algorithm,algorithmicx}
\usepackage[frozencache=true,cachedir=.]{minted}

\usepackage[inline]{enumitem} 
\usepackage{hyperref}

\definecolor{blue}{HTML}{1F77B4}
\definecolor{orange}{HTML}{FF7F0E}
\definecolor{green}{HTML}{2CA02C}

\pgfplotsset{compat=1.14}

\begin{document}
\title{The Mathematics of Comparing Objects}
%
%
\author{Marcus Weber\inst{1} \and
Konstantin Fackeldey\inst{2,1}
}
%
%
\institute{Zuse Institute Berlin (ZIB), Takustra{\ss}e 7, 14195 Berlin, Germany\\
\email{weber@zib.de} \and
TU Berlin, Mathematics, Str. des 17. Juni 136 , 10623 Berlin, Germany\\ 
\email{fackeldey@zib.de}
}
\maketitle              
\begin{abstract}
``After reading two different crime stories, an artificial intelligence concludes that in both stories the police has found the murderer just by random.'' -- To what extend and under which assumptions this is a description of a realistic scenario? 
\end{abstract}
\section{Introduction}

The world is becoming ``more mathematical'' as more and more digital data is collected and analyzed with the aid of computers \cite{Sarker}, but mathematics does not obviously permeate every research area. Comparative text analysis, for example, is far away from being ``solved'' by mathematical algorithms \cite{Da}. The attempt to see literary texts as an input for computational purposes, e.g. as ``vectors'' or as lists of annotations for artificial intelligence, is limited by the complexity of the objects of investigation. However, comparative text analysis is maybe a bad example for mathematical limitations, because texts are already machine-readable codes and therefore in some sense suitable for identifying patterns in these codes. In order to treat general objects of research (e.g. archaeological finds) ``mathematically'' for the purpose of comparison we usually need (i) a specific scientific question, (ii) a predefined annotation of characteristics which is also named ``coordination'' and often (iii) a statistically relevant set of samples. This is especially the case when thinking of mathematics as numerics (which includes machine learning and artificial intelligence). But it is precisely these three aspects that are difficult to implement in day-to-day research. 

\paragraph{Ad (i).} Our preoccupation with research objects is often exploratory, even if we are experts. As we have seen and ``touched'' various objects, an understanding arises in us. As long as our research is exploratory, as long as we do not formulate specific categories, we cannot generate numbers or coordinates for a mathematical analysis. There seems to be a phase in the study of objects that is closed to mathematics -- a phase in which possible patterns do not yet have a name nor a definition. Once we believe to see a pattern or a new relation, we are able to formulate a hypothesis. Now we have a specific research question.  When formulating the research question, the pattern will be given a name (e.g. a certain motif on archaeological vases, a certain lance shape ...). After the hypothesis is formulated, we can ask further questions: Can one falsify the hypothesis? Or do our existing data speak for the correctness of the assumption? We can analyze the context in which the pattern arises. We can (also with the aid of computers) examine the aspects that create or define this pattern. 

\paragraph{Ad (ii).} The mathematization or coordination of objects is done by generating object descriptions in the form of numerical or categorical characteristics (this includes all possible types of data). ``Which characteristics are relevant and how exactly they are to be defined?'' These decisions have to be made before we coordinate the objects. A further decision has to be made about the method or algorithm to be applied to the data. Our analysis results depend on all these decisions. So that ``the mathematics'' does not do the actual research, but only makes our decisions visible in the form of a (visualized) mathematical result which is ``just'' an algorithm-based transformation of the input values, thus, of our decisions. If one wants to be independent of this bias, which occurs through the specification of features, then one would have to be able to execute the mathematical algorithms directly on the research objects without coordination. You would need math beyond numbers or categories. 

\paragraph{Ad (iii).} Research objects (such as literary texts) are generally very complex objects. Compared to the complexity of these objects, the number that we have available for analysis purposes is rather small. It is difficult to substantiate with ``sufficient statistics'' that it is, for example, an important peculiarity that the first words of a specific text are ``Mr. Meyer''. If you only have very few data points available in a high-dimensional data space, then statistical analyzes are generally ``forbidden''. Thus, the statistical analyzes are often preceded by a reduction in complexity. Instead of dealing with the entire research object, the focus is on partial aspects. Furthermore, knowledge can be included in our studies that is located outside the examined object (e.g. the author's family environment or the known functionality of the objects). Such that there is no direct mathematical relation between the research object itself and the actual research result. If such a direct relation does not exist, then also statistical methods (including machine learning) are not able to ``find'' it by just taking the ``measurements'' of the objects into account. Especially this aspect (e.g. trying to derive functionality from form), as well as the aforementioned problems and other biases are also well-analyzed in archaeology \cite{Jung}.  

\begin{figure}[t]
\centering
\includegraphics[width=\textwidth]{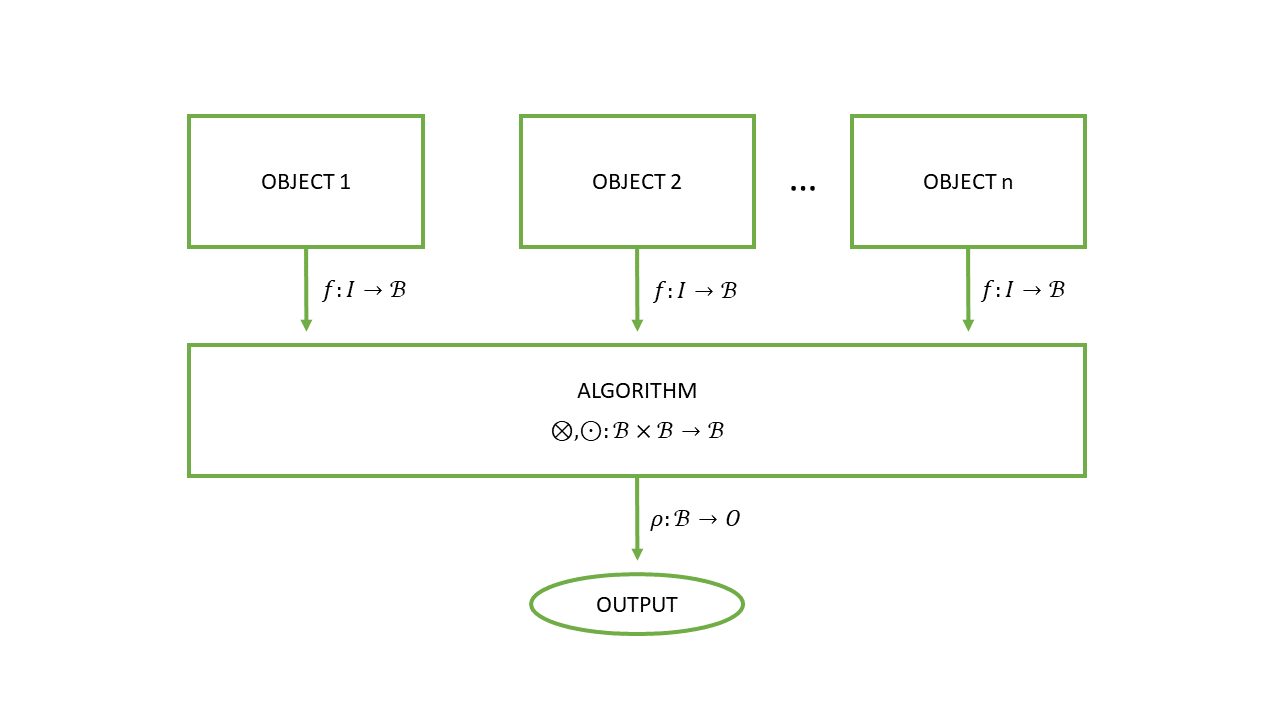}
\caption{\label{fig:origmap}An abstract way of gaining knowledge from comparing objects. The fundamental scheme consists of four different mappings $f,\oplus, \odot,$ and $\rho$.}
\end{figure}

\paragraph{Aim of this article.} Although all the difficulties described above exist, we are able to gain new knowledge and to derive relevant conclusions from comparisons of a small number of very complex objects. There must be a mapping from a list of $n$ objects $(i_1,\ldots, i_n)$, where ``$i$'' stands for {\em input} with $i_k\in I$, to a research result $(i_1,\ldots, i_n)\mapsto o$, where ``$o$'' stands for {\em output} with $o\in O$. Do an experiment! Just watch two movies $i_1$ and $i_2$ and tell me the commonalities $o$ of these two movies. There is this mapping. Depending on the movies and on the interests of the watcher, the answer could look like: ``Both films reproduce a traditional understanding of the role of men''.  The new knowledge is there now. It has been generated in the ``before-math''-phase. Only in ``retrospect'', one could try to build a machine $g$ that can analyze input films and generate  output answers to precisely this research question about the role of men. In fact I could imagine that an artificial intelligence is able to answer the question whether a movie shows certain male roles or not. However, we would need a lot of movies and a more precise definition of ``traditional understanding of the role of men'' for generating training data for this machine. This is the ``math-phase'', but it comes {\em after} we know that the mentioned category is relevant for comparing movies and {\em after} we are able to annotate many movies according to these characteristics. This is not the ``original'' mapping $h(i_1,i_2)=o$. $h$ is the mapping that we did directly after watching the movies. Let us assume, we have created a trained machine to analyze the role of men in movies in the ``math-phase''. From a mathematical point of view, a trained machine represents a mapping $g:I\rightarrow O$ (from one input movie to one answer). The logical way to produce our result concerning the two movies based on $g$ would be: First, apply $g$ to each of these movies and then figure out, whether the output $g(i_1)$ and $g(i_2)$ is ``traditional role of men'' in both cases. In this approach, the act of comparison comes {\em after} the relevant output is produced. Whereas in the original setting the output is generated {\em on the basis} of a comparison. 

In this article it is shown, how the mapping $h: (i_1,\ldots, i_n)\mapsto o$ can be decomposed into four different elementary mappings. These mappings are related to each other which is shown in Fig.~\ref{fig:origmap}. The decomposition into elementary mappings offers the possibility to analyze the nature of $h$ without the {\em need} to provide numerical realizations of the elementary mappings, but giving the {\em opportunity} for a ``partial'' mathematization of the original mapping by replacing some elementary mappings with numerical routines (e.g. with artificial intelligence). 

In a nutshell: Instead of discussing about concrete attempts of coordination of objects, it is discussed how the partial coordination of the intermediate steps of ``gaining knowledge based on a comparison of research objects'' is performed.   

\section{The four elementary mappings}

If we define a mapping like $F:\mathbb{R}^n \rightarrow \mathbb{R}^m$ in the field of numerical mathematics, then this expression, at a first glance, only defines the input and the output domain of $F$. However, we implicitly know many further things:
\begin{itemize}[label=$\bullet$]
    \item We know how the elements of $\mathbb{R}^n$ ``look'' like. In this case, e.g., each element is a list of exactly $n$ real numbers.  
    \item We know how to represent real numbers in a computer. Thus, the elements of $\mathbb{R}^n$ have a clear representation in a numerical algorithm.
    \item We know how to do calculations on the basis of real numbers. We not only know the abstract rules (commutative, associative, distributive law) of the field of real numbers, but we can actually perform calculations for concrete numbers in the computer.
    \item We are often able to write down the rules to perform $F:\mathbb{R}^n \rightarrow \mathbb{R}^m$ such that a computer can be used to compute this mapping for concrete input elements. 
\end{itemize}   
When defining the four elementary mappings in the following sections, then these definitions do not necessarily provide (neither explicitly nor implicitly) any further information of those mentioned types. The definitions of the mappings do not even include necessarily any information about the representation of elements of the input and output domains. In Section \ref{sec:f} and Section \ref{sec:rho} only rather vague definitions of $f$ and $\rho$ in Fig.~\ref{fig:origmap} are presented, while in Section \ref{sec:Bool} the rules for computations are concretely defined.  

\subsection{The mapping $f:I\rightarrow {\cal B}$}\label{sec:f}

The first mapping is $f:I\rightarrow \cal B$. Its input domain is $I$ which includes all possible research objects to be analyzed (could be: ancient text corpora, crime stories, special archaeological objects ...). The set $I$  can also include elements which combine different objects (e.g., a painting together with its frame or together with its Certificate of Authenticity, or other expert knowledge). The elements of this set are far away from being numbers or from being coordinated. $f$ maps the elements of $I$ to an output set $\cal B$. In order to understand $\cal B$ we first have to explain another set $T$. This is the set of all possible thoughts. Maybe it is easier to think of ``written thoughts'', i.e. of texts. Then $T$ would be the set of texts -- texts that have already been written, texts that will be written or {\em can} be written. 

Now $f(i)$ is a {\em subset} of $T$, which includes everything that can be thought about the object $i\in I$. This definition of $f$ also offers the possibility to allow for different  contradicting opinions about the research object, or thoughts that arise from other sources of knowledge, or to explicitly express doubts, or to formulate conditional thoughts that would depend on a yet unknown fact. 

With this definition, $f(i)$ can be written as an {\em element} of ${\cal B}$, where ${\cal B}={\cal P}(T)$ is the power set (the set of all subsets) of $T$. The element $0\in{\cal B}$ denotes the empty set. In this sense, via $f:I \rightarrow {\cal B}$ every object $i$ is mapped to the set of possible ``thoughts'' about $i$, denoted as $f(i)$, see also \cite{Weber}. 

At this point it becomes clear that the definition of $f$ really does not make any statements about how the elements of the sets $I$ and $\cal B$ can be represented -- and for sure not about their representation in a computer. It is also not said, how this mapping can be ``calculated''. The fact that there is a mapping $f:I\rightarrow{\cal B}$ actually only indicates its input and its output domains. It should not be said that for the sets $I$ and $\cal B$ the representation in a computer has to be taken into account immediately. In the following there will also be examples of how these constructs can be used outside of computing machines. Because of this generality, the function $f$ can also be seen as a ``pre-math'' object. Assuming that $T$ is an infinite set, then $\cal B$ is for sure an uncountably infinite set due to Cantor \cite{Cantor}. He proved that the power set of an infinite set $T$ has a higher cardinality than $T$. Computers can only represent finite sets of elements. This could be a counter argument against the hope to operate with a construction like $\cal B$ in a computer. 

However, computers are constructed to represent real numbers, at least somehow a subset of them. The set of real numbers is also uncountably infinite. Real numbers are represented in the computer by a finite string of zeros and ones.  Interestingly, these strings would be a possible and also a perfect representation of elements of $\cal B$, if the set $T$ is assumed to be finite. In this case, every bit of the string corresponds to one element of $T$. By definition,  $A\in \cal B$ is a subset of $T$. It can in fact be represented by the string of bits. Whenever the respective value of the bit in the string is one, then the corresponding element of $T$ belongs to $A$. If the bit has value zero, then the respective element of $T$ does not belong to the subset $A$. The elements of $\cal B$ are then represented by strings of bits.

Here it should only be argued that there might be a way to represent elements of $\cal B$ in a computing machine. It is not said to which concrete elements of $T$ -- to which concrete thoughts -- the bits are assigned or even which finite set of thoughts should form $T$. It is not our aim to say, that representing elements of $\cal B$ on the basis of bits must include a one-to-one correspondence between elements of $T$ and those bits. 

Thinking of representing the set $\cal B$ somehow by strings of bits: How far does this definition of a coordination like $f$ differ from what usually digital or numerical algorithms apply? Representing objects in a computer {\em always} ends up in a string of bits. If we furthermore aim at constructing an algorithm that makes conclusions just on the basis of these strings of bits which is an aim of artificial intelligence, then even the {\em meaning} of $f(i)$ is equivalent. Everything that can be thought about the object is included in its string of bits, which in turn is thus also a representation of $\cal B$.

\subsection{The mapping $\rho:{\cal B}\rightarrow O$}\label{sec:rho}

The set of all considerations $f(i)\in \cal B$ that can be thought about a particular object $i\in I$ has been explained in the last section. Some thoughts about objects can be trivial: ``The detective in this crime story wears clothes and sits on a chair. In the crime story, the sun shines during the day and the moon is seen in the evening, unless it is cloudy or there is a new moon or a lunar eclipse, or you are looking into the wrong direction.'' 

Most thoughts are not rich enough to serve as an output of a scientific interpretation. The role of the mapping $\rho:{\cal B}\rightarrow O$ is to extract meaningful thoughts $o\in O$ from the set of all possible thoughts $A\in \cal B$. The role of the mapping $\rho$ may not be difficult to understand, but such mappings are difficult to be algorithmized on a computer. The algorithms of artificial intelligence, e.g., can be seen as executing the functions $f$ and $\rho$ one after the other in a specific way. The sequence $I\rightarrow {\cal B} \rightarrow O$ can thus also model the process of an AI analysis: First, the object $i\in I$ is digitized and represented by $f(i)\in \cal B$, then a meaningful conclusion is derived from this digitized form $g(i)=\rho(f(i))\in O$: ``The movie reproduces a traditional understanding of the role of men.'' The mapping $\rho$ presents its output in a human-understandable way. In this regard the output $o\in O$ could be a statement, a number, or more complex like a visualization of a network.

If there were a clear distinction between the meaningful and the meaningless thoughts, then $o=\rho(A)$ would ``simply'' be a filter that eliminates the meaningless thoughts from the set $A\in \cal B$. The meaning of a thought, however, does not lie in the thought itself, but in its ability to express something special or something regular. This implies that relevant thoughts arise from comparing objects (which includes abstract research objects). In this regard, even wearing a trench coat can become an important aspect of considerations about crime movies: ``Peter Falk, aka Columbo, wore the same trench coat until 1978. It didn't come from the film prop, but from Falk's private wardrobe.'' This statement is made by comparing the individual episodes of the Columbo series. The statement takes knowledge into account which is not visible within the movies.  

Of course, while watching a single episode one could have had the thought that the actor is wearing his own trench coat, but the {\em relevance} of this thought comes about through the recurrence of this observation in the other episodes. In this manuscript, the mappings $f:I\rightarrow \cal B$ and $\rho:{\cal B}\rightarrow O$ are therefore not carried out directly one after the other, but the act of comparing objects is interposed, see Fig.~\ref{fig:origmap}. In this way, the act of comparison becomes a part of the filtering process leading to relevant conclusions. The mapping $\rho:{\cal B}\rightarrow O$ is then still a filtering or a complexity reduction, but its role for ``translating'' the resulting set $A\in \cal B$ into a human-understandable output is emphasized. 

\subsection{The mappings $\oplus,\odot:{\cal B}\times{\cal B}\rightarrow {\cal B}$}\label{sec:Bool}

Although the sets $I, O,$ and $\cal B$ have been defined in the previous sections, the indeterminacy of the representation of the elements of $\cal B$ makes the two-step-approach $I \rightarrow {\cal B} \rightarrow O$ almost meaningless for the generation of an output $o=g(i)=\rho(f(i))$ on the basis of an input object $i$.  
On the one hand, there is a good reason -- according to what has been said so far -- for the setting ${\cal B}=I$ and considering $f$ as identity function. In this case, the objects themselves represent everything that can be thought about them. The ``whole work'' to filter out something meaningful is then done by the mapping $\rho$. On the other hand, also the opposite extreme ${\cal B}=O$ is reasonable with $\rho$ being the identity function. In this case, $O$ (e.g.,  written deep analyzes of the objects) represents everything that can be thought about the objects and the ``whole work'' of identifying these representatives is done by the mapping $f$. 

The uncertainty about the set ${\cal B}$ comes about because we have allowed an extreme range of possible representations in Section \ref{sec:f}. In the followings we will again not define the representation of the elements of $\cal B$, but we will see how one must define operations on $\cal B$ so that the act of comparison can be implemented in between $f$ and $\rho$. By defining operations within $\cal B$, the nature of this intermediate set becomes clear.

According to the Section~\ref{sec:rho} the act of comparison is used to filter out or reduce the sets of possible thoughts about the objects. Given all possible thoughts of all considered objects as input ${\cal B}\times\ldots\times{\cal B}$, the act of comparison selects a subset of these thoughts, i.e., the output is again from the set $\cal B$, see Fig.~\ref{fig:origmap}. However, the resulting subset is not arbitrary, but it is based on real delimitation processes between the objects. The subset can thus be described in the form of differences and commonalities of the entered objects: ``The commonality of all thoughts about the episodes of Columbo up to 1978, but none of the thoughts about the later episodes''.  Here, the resulting set $A \in \cal B$ would contain the thoughts about the trench coat. 

From set theory it is known that such subsets can be represented solely with the aid of two operations: intersection $\odot$ and symmetrical set difference $\oplus$. In this regard $({\cal B}, \oplus, \odot)$ is a Boolean ring. The following considerations have been adapted from \cite{Weber}.  A Boolean ring shares some rules of computation with real numbers, because both are algebraic rings: 

\begin{definition} \label{def:ring}
Given a set $\cal B$ and two binary operations $\oplus:{\cal B}\times {\cal B}\rightarrow {\cal B}$ and $\odot: {\cal B}\times {\cal B}\rightarrow {\cal B}$. Then $({\cal B},\oplus, \odot)$ is denoted as an {\em algebraic ring}, if the following conditions hold for all (not necessarily pairwise different) elements $A,B,C\in {\cal B}$:
\begin{itemize}
    \item[(i)] the laws of distribution: $A\odot(B\oplus C)=(A\odot B) \oplus (A\odot C)$ and\\ $(B\oplus C)\odot A=(B\odot A) \oplus (C\odot A)$,
    \item[(ii)] the associative law: $(A\odot B)\odot C=A\odot(B \odot C)$, and
    \item[(iii)] that $({\cal B},\oplus)$ is a commutative group, i.e., 
    \begin{itemize}
        \item[a)] the associative law holds: $(A\oplus B) \oplus C=A\oplus (B \oplus C)$,
        \item[b)] commutivity holds: $A \oplus B=B\oplus A$, 
        \item[c)] there exists an element $0\in\cal B$, such that $0\oplus A=A$ for all $A\in \cal B$, and
        \item[d)] for every $A\in\cal B$ there is an element $\overline{A}\in \cal B$ such that $A\oplus \overline{A}=0$.
    \end{itemize}
\end{itemize}
\end{definition}

There is one important further equation for $\cal B$. It is the idempotency $A \odot A= A$ which additionally holds and which turns the algebraic ring into a Boolean ring. By this property the algebra of real numbers differs from the algebra of $\cal B$: 

\begin{definition} \label{def:Bool}
An algebraic ring $({\cal B}, \oplus, \odot)$ is denoted as {\em Boolean ring}, if idempotency $A \odot A= A$ holds for every $A\in {\cal B}$.
\end{definition}

The definition of an algebraic ring or of a Boolean ring does not include a neutral element of multiplication, i.e., we not necessarily have to assume an element $1\in {\cal B}$ with $1\odot A=A$ for all $A\in{\cal B}$. A ring which has such an element $1$ is called a {\em ring with unity}. In our case, the complete set of thoughts $T\in{\cal B}$ has this role, i.e., $1=T$. In this sense, the expression $B=1\oplus A$ means, that we create the subset $B$ of all thoughts which are {\em not} element of $A$. $B$ is the complement of $A$. 

Boolean rings (and Boolean algebras) are studied in complexity analysis, computational algebra, and in computer science. The arithmetic laws formulated in the two definitions can be used to transform equations. From idempotency some further properties of Boolean rings can directly be derived. For instance, the equation $A\oplus A=0$ formalizes that there is nothing to be written when we want to figure out the differences between $A$ and $A$. In other words the element $\overline{A}$ in item (iii d) of the definition of an algebraic ring is equal to $A$. The equation $A\oplus A=0$ does not occur in the definition of a Boolean ring, because it is already a consequence of $A\odot A=A$.  This can be shown in the following way: $(A\oplus A)\odot(A\oplus A)=A\oplus A$ by the idempotency. Furthermore, $(A\oplus A)\odot(A\oplus A)=A\oplus A\oplus A\oplus A$ by the law of distribution. Thus, $A\oplus A\oplus A\oplus A=A\oplus A$, which shows $A\oplus A=0$.  

Also the commutative law $A\odot B=B\odot A$ is a consequence of $A\odot A=A$ and of $A\oplus A=0$. Note, that $A\oplus B = (A \oplus B)^2=A\oplus (A\odot B) \oplus (B\odot A) \oplus B$. This means $(A\odot B) \oplus (B\odot A)=0$, which proves the commutative law of multiplication.  Boolean rings are commutative rings. 

\paragraph{Nontrivial intermediate.} The intermediate step between ``coding'' the objects via $f$ and ``understandable interpretation'' via $\rho$ is performed by an algorithm based on operations $\oplus$ and $\odot$ in $\cal B$. The algorithm can be a branched algorithm, the branching conditions of which depend on evaluations of intermediate steps. How exactly the branching functions are to be set up is not discussed here. It is only intended to show that the act of comparison can determine very complex relationships, but all of them are based on the determination of commonalities or differences between the input sets in $\cal B$. Why does this approach lead to a nontrivial intermediate set $\cal B$ in the sequence $I\rightarrow {\cal B} \rightarrow O$? The reason for ${\cal B}\not=I$ is given by the fact that the mappings are usually not definable as $\odot,\oplus: I\times I\rightarrow I$. The reason for ${\cal B}\not=O$ is given by the fact, that $O$ does not include everything that can be thought about the objects. Is there at least somehow a possibility to use computers to calculate $\oplus$ and $\odot$? Going back to one specific representation of $\cal B$ in Sec.~\ref{sec:f} which has been using strings of bits with a one-to-one correspondence between the elements of a finite set $T$ and the bits. In this special case, the operations $\oplus$ and $\odot$ are just performed by bit-wise applications of the logical XOR or AND functions to the strings of bits. However, in the very general setting, this bit-wise approach is not the only possibility.

\section{Exemplified realizations of the fundamental scheme}

Coding ($f$), performing calculations $(+,\cdot)$ and interpreting them ($\rho$) are actually common to all numerical analyses of non-numerical coordinated objects. In this chapter the emphasis should be placed on the research phase in which the coordination of the objects has not yet been completed. Calculations are based on $\oplus$ and $\odot$. Three different scenarios of {\em a priori} information are conceivable. The following sections provide an example for each of these levels. 
\begin{itemize}
    \item {\bf Characteristics are not yet available.} Simplified assumptions about the objects and their differences exist, maybe from metadata about them. This leads to an algorithm which extracts the relevant list of characteristics from comparisons of objects, see Sec.~\ref{sec:algo}.  
    \item {\bf A list of the characteristics of the objects is not yet available, but objects can be studied by an expert.} The approach aims at a meaningful clustering of objects such that the relevant characteristics can be identified {\em a posteriori} from the cluster assignments, see Sec.~\ref{sec:ca}.
    \item {\bf For the objects a list of assigned characteristics is available.} In a kind of feature extraction, those are to be selected from a given list of possible features that are most important for a specific result, see Sec.~\ref{sec:fs}. This can also be done with an algebraic approach (algebraic feature extraction) which is related to Formal Concept Analysis, see Sec.~\ref{sec:afs}. 

\end{itemize}

\subsection{Generating an algorithm}\label{sec:algo}
Although the set $T$ of possible thoughts might be infinite, the number of objects of
investigation is usually finite. Imagine {\em all possible} algorithms that can be applied on a finite set of input {\em real} numbers. Although the finite set of real numbers to be used in the algorithm is assumed to be pre-defined, the set of possible output numbers is not finite and depends on the algorithms. Real numbers and elements of Boolean rings share the properties of Def.~\ref{def:ring}. 
However, idempotency has deep consequences for the structure and output of algorithms. 

Let us assume, that we only deal with a finite set of objects $i_1,\cdots, i_m$, where the number of objects $m$ and the number $n$ of input elements of the algorithm in Fig.~\ref{fig:origmap} can be different.
If the set of possible input elements $f(i_1), \cdots, f(i_m)$ is finite, then only $2^{{2^m}-1}$ output values of an algorithm in $\cal B$ are possible. Although we can imagine an infinite set of different algorithms (with different numbers $n$ of input slots) to be performed on a finite set of objects, there is only a finite number of possible outcomes. This is a speciality of idempotency. The number $2^{{2^m}-1}$ comes from the fact, that due to the distributive laws in Def.~\ref{def:ring}, every possible output can be written as a sum $\oplus$ of products $\odot$ of elements $f(i_1), \cdots, f(i_m) \in \cal B$. Due to $A\oplus A=0$, every possible product expression can only occur at most once in this sum. Due to $A\odot A=A$, every possible factor can also only occur at most once in each product expression.

\paragraph{Example.} Assume, we analyze $2$ objects with $A=f(i_1)$ and $B=f(i_2)$. Then there are $8$ possible outcomes presented in Tab.~\ref{tab:my_label1}.

\begin{table}[ht]
    \vspace*{0.2cm}
    \centering
    \begin{tabular}{|c|c|c|}
        \hline
        0 &   the empty set & 0\\
        $A$ & everything about object $i_1$& 1\\
        $B$ & everything about object $i_2$& 2\\
        $A\odot B$ & all commonalities&3 \\
        $A\oplus B$ & all differences&4\\
        $A\oplus (A \odot B)$& $i_1$ has it, but $i_2$ has not&5\\
        $B\oplus (A \odot B)$& $i_2$ has it, but $i_1$ has not&6\\
        $A\oplus B \oplus (A \odot B)$ & union of $A$ and $B$&7 \\
        \hline
    \end{tabular}
    \vspace*{0.2cm}
    \caption{Possible outputs of algorithms based on 2 objects.}
    \label{tab:my_label1}
\end{table}

Using this table of all possible algebraic terms based on two elements $A$ and $B$, we end up with a Boolean ring consisting of $8$ elements. In Tab.~\ref{tab:my_label2} the operations $\oplus$ and $\odot$ for these $8$ elements are shown. The elements of a Boolean ring can also be ordered partially. Using the subset property or the definition $Y\geq X \Leftrightarrow X=Y\odot X$, this partial order is constructed in Tab.~\ref{tab:my_label2} on the right. 

\begin{table}[bh]
\vspace*{0.2cm}
    \centering
    \begin{tabular}{c|cccccccc}
         $\oplus$ & 0 & 1 & 2 & 3&  4 & 5 &6  &7  \\
         \hline
         0 & 0 & 1 & 2 & 3 & 4 & 5 & 6 & 7 \\
         1 & 1 & 0 & 4 & 5 &2 & 3&7&6\\
         2& 2& 4 & 0& 6 & 1& 7 & 3 &5\\
         3& 3 & 5 & 6& 0 & 7& 1 &2 &4\\
         4& 4& 2 & 1& 7& 0& 6 &5 & 3\\
         5& 5& 3& 7& 1& 6& 0& 4 & 2\\
         6& 6 & 7& 3& 2& 5& 4& 0& 1\\
         7& 7 & 6& 5& 4& 3& 2& 1 &0
    \end{tabular} 
    \hspace*{1cm}
    \begin{tabular}{c|cccccccc}
         $\odot$&  0 & 1 & 2 & 3 & 4 & 5 & 6 & 7 \\
         \hline
         0&  0 & 0 & 0 & 0 & 0 & 0 & 0 & 0 \\
         1&  0 & 1 & 3 & 3 & 5 & 5 & {\bf 0} & 1 \\
         2&  0 & 3 & 2 & 3 & 6 & {\bf 0} & 6 & 2 \\
         3&  0 & 3 & 3 & 3 & {\bf 0} & {\bf 0} & {\bf 0} & 3 \\
         4&  0 & 5 & 6 & {\bf 0} & 4 & 5 & 6 & 4 \\
         5&  0 & 5 & {\bf 0} & {\bf 0} & 5 & 5 & {\bf 0} & 5 \\
         6&  0 & {\bf 0} & 6 & {\bf 0} & 6 & {\bf 0} & 6 & 6 \\
         7&  0 & 1 & 2 & 3 & 4 & 5 & 6 & 7 \\
    \end{tabular}
    \hspace*{1cm}
    \begin{tabular}{c|cccccccc}
         $\geq$&  0 & 1 & 2 & 3 & 4 & 5 & 6 & 7 \\
         \hline
         0&  + &  &  &  &  &  &  &  \\
         1&  + & + &  & + &  & + &  &  \\
         2&  + &  & + & + &  &  & + &  \\
         3&  + &  &  & + &  &  &  &  \\
         4&  + &  &  &  & + & + & + &  \\
         5&  + &  &  & &  & + &  &  \\
         6&  + &  &  &  &  &  & + &  \\
         7&  + & + & + & + & + & + & + & + \\
    \end{tabular}
    \vspace*{0.2cm}
    \caption{Operations with elements of a Boolean ring. The numbering is defined in the last column of Tab.~\ref{tab:my_label1}. Using the algebraic terms in the fist column of Tab.~\ref{tab:my_label1}, the tables for $\oplus$ and for $\odot$ have been created. $\oplus$ forms a commutative group. $\odot$ has non-trivial dividers of zero (marked in bold). Using the $\odot$-table and the definition $Y\geq X \Leftrightarrow X=Y\odot X$, the last table has been created. From this table one can read, e.g., that $4\geq 5$ is a valid expression. }
    \label{tab:my_label2}
\end{table}

Just by knowing the number $m$ of {\em basic objects}, one can construct a partially ordered Boolean ring $\cal B$ with unity having $2^{2^m-1}$ elements. If the number of basic objects is $2$, then $f$ is a mapping from $I$ to the set $\{0,1,2,3,4,5,6,7\}=\cal B$ and $\rho$ is a mapping from $\{0,1,2,3,4,5,6,7\}$ to a human-understandable output. $\cal B$ is {\em complete} in the sense, that it includes all elements that can ``occur within algorithms''. The number $7$ represents the unity of multiplication in the given example (have a look at the last column and row of the $\odot$-table). 

\paragraph{Why {\em basic} objects?} In the last example it seams that $m$ should be the number of objects and $f(i_k)=k$ for $k=1,\ldots,m$, but this is not necessarily meant.  Why have $A$ and $B$ been denoted as {\em basic} objects? There is a reason for this. When we learned categories as we were children, then we did this on the basis of comparisons. We had to look at a lot of pictures with blue objects until we understood the meaning of ``blue''. The meaning became clear, when our parents told us, that ``blue'' is the commonality of certain objects or it is a characteristic that differentiates between them. The objects that have been used to explain categories to us (this is a saga and that is a heroic epic) are the {\em basic objects}. Imagine three basic objects: $A=$ ``big red triangle'',  $B=$ ``small red circle'', and $C=$ ``small blue triangle''. We already know that a Boolean ring based on these three basic objects has 128 elements. Among these elements there are also the characteristics like shown in Tab.~\ref{tab:my_label3}.

\begin{table}[h]
    \vspace*{0.2cm}
    \centering
    \begin{tabular}{|c|c|}
         \hline 
         big &  $A\oplus (B\odot A)\oplus (C\odot A)\oplus(A\odot B \odot C)$\\
         small &  $(B\odot C)\oplus (A\odot B \odot C)$\\
         circle &  $B\oplus (A\odot B)\oplus (C\odot B)\oplus(A\odot B \odot C)$\\
         triangle & $(A\odot C)\oplus(A\odot B \odot C)$ \\
         blue & $C\oplus (A\odot C)\oplus (B\odot C)\oplus(A\odot B \odot C)$ \\
         red &  $(A\odot B)\oplus(A\odot B \odot C)$\\
         \hline
    \end{tabular}
    \vspace*{0.2cm}
    \caption{How the six characteristics are represented by the three basic objects. They are included in the Boolean ring with $2^7=128$ elements.}
    \label{tab:my_label3}
\end{table}

Using these characteristics, other objects are also represented by elements of the Boolean ring. A ``small blue circle'' would be represented by the union of the three corresponding characteristics, where the union of two elements $X$ and $Y$ can be expressed via $X\oplus Y\oplus (X\odot Y)$. Or simpler: $X\oplus Y\oplus XY$. Starting with the union of ``small'' and ``blue'', we can simplify:
\begin{eqnarray*}
 && (BC\oplus ABC) \oplus (C \oplus AC \oplus BC\oplus ABC) \oplus \\
 && (BC\oplus ABC)(C\oplus AC\oplus BC\oplus ABC)\\ &=& C \oplus AC. 
\end{eqnarray*}
Adding ``circle'' to this expression finally leads to:
\begin{eqnarray*}
&& (C\oplus AC) \oplus (B\oplus AB \oplus BC\oplus ABC) \oplus (C \oplus AC)(B\oplus AB \oplus BC\oplus ABC) \\
&=& C\oplus B\oplus AC \oplus AB\oplus BC\oplus ABC.
\end{eqnarray*}
The expression $C\oplus B\oplus AC \oplus AB\oplus BC\oplus ABC$ represents the ``small blue circle''. The characteristics as well as the objects (which can be described by these characteristics) are elements of the same set $\cal B$. The interesting thing about the elements in Tab.~\ref{tab:my_label3} and about $A\odot B\odot C$ is, that these elements $X$ do not allow for relations $X\geq Y$ except for the trivial ones, $X\geq 0$ and $X\geq X$, for any ring element $Y$. These elements are not supersets of any other elements. Thus, they are the most detailed characteristics of the mentioned objects. They are also denoted as {\em atoms} in Boolean theory.

\paragraph{The difference between objects and characteristics.} The role of $f$ is to assign an element of $\cal B$ to an object. The role of an algorithm is to extract characteristics. Elements of $\cal B$ which are supersets of many other elements are suitable to represent objects. Elements which are only supersets of a small number of elements are suitable to represent characteristics. In the above example with  $8$ different elements, one can count the $\geq$-relations per row in Tab.~\ref{tab:my_label2} on the right. Here, the elements 7,1,2, and 4 are more suitable for objects, while 3,5, and 6 may better represent characteristics. 

\paragraph{Algebraic equations.} Consider a new example with three objects $a, b,$ and $c$. Some logical considerations about the three objects can be formulated in terms of algebraic equations. For example, if we want to focus on cultural differences on the basis of archaeological findings (we want to restrict our thoughts $T$ accordingly, such that commonalities of cultures are not expressed) and if we know that $a$ and $c$ stem from different cultures, then $a\odot c=0$ might model this assumption. Furthermore, assume that $b$ stems from the same culture like $c$, but from an earlier period. In this case, the assumption that $c$ is very similar to $b$ but has a higher complexity, leads to the inequality $c\geq b$. Relations between the objects (or between comparisons of the objects) can be formulated in terms of algebraic equations. Also the inequality $c\geq b$ can be written as $b\oplus (c\odot b)=0$. From these two equations ($a\odot c=0$ and $b\oplus (c\odot b)=0$) further equations can be derived, like $a\odot b =0$ or $a\odot b\odot c=0$. In fact, all expressions that are element of the ideal created by $a\odot c$ and $b\oplus (c\odot b)$ are assumed to be zero. The interesting point here is, that Boolean rings are principle ideal rings. This means that ideals are always generated by just one expression. The ideal is always generated by an element, which is the ``union'' of the creating expressions. The union of two expressions $X$ and $Y$ is computed via $X\oplus Y\oplus (X\odot Y)$.  In our case, we arrive at an ideal which is generated by $(a\odot c)\oplus b \oplus (c\odot b)$. Since the ideal is {\em generated} by this expression, $(a\odot c)\oplus b \oplus (c\odot b)=0$ is the only condition to be checked, in order to analyze our assumptions. Instead of checking many equations, it is enough to compute the generator of the principal ideal and to only check one condition. In order to evaluate $(a\odot c)\oplus b \oplus (c\odot b)$, one could rewrite it into $((a\oplus b)\odot c)\oplus b$: First one has to write down all differences between $a$ and $b$, then one has to extract from this list only those characteristics which are in common with object $c$, and then one has to check, whether in this remaining list there are characteristics which are different from the characteristics of $b$. If we want to restrict our thoughts according to our assumptions about the objects, then all characteristics included in this difference list are irrelevant. 

In Sec.~\ref{sec:afs} it is discussed, how these principle ideals can be generated on the basis of given data.   

\paragraph{Reduction of complexity.} Reducing the complexity is a main tool in studying complex objects. We can reduce complexity by reducing the number of basic objects. Thus, we can try to assign elements of the 8-elements-ring $\cal B$ to the three mentioned objects $a,b,$ and $c$. This ring is based on only two basic objects and, thus, does not model complex situations. The following example is therefore very simple. A possible solution of the algebraic system is $a=6, b=5, c=1$, because $((a\oplus b)\odot c)\oplus b=((6\oplus 5)\odot 1)\oplus 5=0$. The role of an algorithm is to extract characteristics on the basis of the input objects. In the very simple case of the 8-elements-ring $\cal B$ the elements $3,5,6$ are characteristics. The characteristics $a=6$ and $b=5$ can be extracted, by analyzing and describing the objects $a$ and $b$. By looking into Tab.~\ref{tab:my_label2} one can find an algorithm to extract $3$. The corresponding formula is $b\oplus c$. By describing the difference between $b$ and $c$, we get the missing characteristic $3=5\oplus 1$. The basis of an algorithm is a formula which describes, what kind of comparisons are needed in order to extract characteristics. Writing down the human-understandable result of such a comparison is the mapping $\rho$. According to our assumptions, the objects $a$ and $b$ should be described and the difference $b\oplus c$ should be evaluated. In these three descriptions, the irrelevant characteristics $((a\oplus b)\odot c)\oplus b$ should be deleted. This procedure provides the most detailed characteristics of the three objects with regard to our assumptions.

\paragraph{Finding the algorithm.} Complexity reduction has been presented as the idea to restrict the studies to a small number of basic objects and to find a solution of the algebraic system on the set of elements of a Boolean ring (trying to identify objects with elements which have a lot of $\geq$-relations). However, there is a very systematic way to proceed with given algebraic equations on the input objects. Algebraic equations generate a principle ideal in the Boolean ring $\overline{\cal B}$ taking all input objects as basic objects. $\overline{\cal B}$ has three basic objects $a,b,c$ in our example, thus, 128 elements. In our example above, the equations $a\odot c=0$ and $b\oplus(b\odot c)=0$ generate a principle ideal $\langle ((a\oplus b)\odot c)\oplus b \rangle \subset \overline{\cal B}$. Now two sets of expressions play an important role. The first set is the ideal. In this case it has 16 elements (written in short form):
\begin{eqnarray*}
\langle ((a\oplus b)\odot c)\oplus b \rangle &=&\{
 0, {\bf ac}, ab, abc, {\bf b\oplus bc}, ab\oplus abc, ab \oplus ac, ac \oplus abc,\\
&& ac\oplus b\oplus bc, ac \oplus ab \oplus abc, ab \oplus b \oplus bc, \\
&& abc \oplus b \oplus bc, ac \oplus b \oplus bc \oplus ab, ac \oplus b \oplus bc \oplus abc,\\
&& ab \oplus b \oplus bc \oplus abc, ac \oplus ab \oplus abc \oplus b\oplus bc\}.
\end{eqnarray*}
The other set ${\cal C}\subset \overline{\cal B}$ consists of zero dividers representing the characteristics. The following 8 elements of $\overline{\cal B}$ only allow for trivial $\geq$-relations (they can also be read from Tab.~\ref{tab:my_label3}). Every product of two different elements of $\cal C$ is zero: 
\begin{eqnarray*}
{\cal C}&=& \{ 0, abc, a\oplus ab\oplus ac\oplus abc, bc\oplus abc, b\oplus ab\oplus bc\oplus abc, ac\oplus abc,\\
&& c \oplus ac \oplus bc \oplus abc, ab \oplus abc\}.
\end{eqnarray*}

In order to find the expressions which provide relevant characteristics one has to map every element of $\cal C$ to its residue modulo $\langle (a\odot c)\oplus b \oplus (c\odot b) \rangle$:
\begin{eqnarray*}
0 &\mapsto& 0\\
abc &\mapsto& 0\\
a\oplus ab\oplus ac\oplus abc &\mapsto& {\bf a}\\
bc \oplus abc &\mapsto& {\bf b} \\
b \oplus ab\oplus bc \oplus abc &\mapsto& 0 \\
ac\oplus abc &\mapsto& 0\\
c \oplus ac\oplus bc \oplus abc &\mapsto& {\bf b \oplus c}\\
ab \oplus abc &\mapsto& 0.
\end{eqnarray*}
The output set of this mapping provides the formulas to compute all relevant characteristics: $\{a, b, b \oplus c\}$. Generating the corresponding descriptions and deleting the list of irrelevant characteristics $((a\oplus b)\odot c)\oplus b$ is the proposed method.

\paragraph{Possibility of cooperation.} The algebraic formulation of the necessary comparisons that have to be carried out in order to achieve a certain result offers the possibility of having experts with separate areas of knowledge work together. Imagine you want to extract the peculiarities of four different texts, i.e. what properties the individual texts have that the other texts do not have. The special properties of the texts $a, b, c,$ and $d$ can be represented by the following four expressions
\begin{eqnarray*}
&&a\odot (1\oplus b)\odot (1\oplus c)\odot (1\oplus d),\cr
&&b\odot (1\oplus a)\odot (1\oplus c)\odot (1\oplus d),\cr
&&c\odot (1\oplus a)\odot (1\oplus b)\odot (1\oplus d),\cr
&&d\odot (1\oplus a)\odot (1\oplus b)\odot (1\oplus c),
\end{eqnarray*}
where e.g. $(1 \oplus a)$ is the complement of $a$. Taking the union of these four expressions and expending the formula leads to: $abc \oplus abd \oplus acd \oplus bcd \oplus a\oplus b\oplus c\oplus d$. This is the algebraic term to be ``computed'' when extracting the whole list of specialities of the four texts. 

Now imagine there are two experts. One expert only knows very well texts $a$ and $b$, whereas the other expert only knows $c$ and $d$ very well. By rewriting the formula as $((1\oplus cd)\odot (a \oplus b))\oplus ((1\oplus ab)\odot (c\oplus d))$, we can arrange the cooperation of the two experts in the following way:
\begin{itemize}
    \item Assumption: It is easier to find all commonalities than all differences. Thus, the experts determine all common characteristics of their own two texts. This provides $x=ab$ and $y=cd$. Then the experts exchange this knowledge.
    \item The expert for the texts $a$ and $b$ now determines all differences $(a\oplus b)$ which are not in the ``list'' $y$, i.e., $u=(1\oplus y)\odot (a\oplus b)$. Every common characteristic of $c$ and $d$ which is not a commonality of $a$ and $b$ is already a peculiarity. In the same way: $v=(1\oplus x)\odot(c\oplus d)$.
    \item In the end the experts again exchange knowledge and write down all differences between these two ``lists'' $u$ and $v$, i.e. the final result is given by $u\oplus v$.
\end{itemize}

\subsection{Sorting objects}\label{sec:ca}

The next realization is inspired by an analysis of kernels of graph Laplacians. Imagine a Boolean ring with a partial ordering, like in Tab.~\ref{tab:my_label2} on the right. The $\geq$-table can be regarded as an adjacency matrix of a directed graph. Directed edges between two different elements of the Boolean ring (the vertices) occur if there is a $\geq$-relation between these elements. Here we account for all elements except for the empty set $0$. This matrix is transformed into a graph Laplacian $\bf L$ by adjusting the diagonal elements in such a way that the row sums of this matrix are zero. In the given example:
\[
{\bf L}=
\begin{pmatrix}
 -2 & 0 & 1 & 0 & 1 & 0 & 0\cr
  0 & -2& 1 & 0 & 0 & 1 & 0\cr
  0 & 0 & -0 & 0 & 0 & 0 & 0\cr
  0 & 0 & 0 & -2 & 1 & 1 & 0\cr
  0 & 0 & 0 & 0 & -0 & 0 & 0\cr
  0 & 0 & 0 & 0 & 0 & -0 & 0\cr
  1 & 1 & 1 & 1 & 1 & 1 & -6
\end{pmatrix}
.
\]
The off-diagonal entries ($1$ and $0$) are interpretable as transition rates between the vertices. A Markov process is defined in that way. The elements that have been denoted as ``characteristics'' are then like ``sinks'' (or reaches) of the corresponding process (all rates are zero in the corresponding rows). It is known that these graph Laplacians have an $n$-dimensional kernel, where $n$ is the number of ``characteristics'' (i.e., of reaches)\cite{kernel}. A corresponding basis of this kernel consists of non-negative vectors which can be regarded as ``committor functions'' of the process \cite{metzner}. In general, each of these vectors has the following structure: The entry of the vector is $1$ for exactly one of the characteristics. The entry is $0$ for all Boolean ring elements which do not share commonalities with this characteristic ${ c}$. Elements of the Boolean ring which are supersets of the corresponding characteristics ${ c}$ have values in between $0$ and $1$. The exact value corresponds to the probability to end up in characteristic $ c$ when ``starting'' the process in the given ``object'' and depends on how much the characteristic $ c$ already represents the given Boolean ring element (are there many subset relations?). In our example we get 3 eigenvectors for the three characteristics $c\in\{3, 5, 6\}$:
\[
\left(\begin{array}{c}
 1/2\cr
 1/2\cr
  1 \cr
  0 \cr
  0 \cr
  0 \cr
  1/3
\end{array}
\right)
,
\left(\begin{array}{c}
 1/2\cr
  0\cr
  0 \cr
 1/2 \cr
  1 \cr
  0 \cr
  1/3
\end{array}
\right)
,
\left(\begin{array}{c}
  0\cr
 1/2\cr
  0\cr
 1/2 \cr
  0 \cr
  1 \cr
 1/3
\end{array}
\right)
\quad
\left.\begin{array}{l}
  A\cr
  B\cr
  A\odot B\cr
  A\oplus B \cr
  A\oplus (A\odot B) \cr
  B\oplus (A\odot B) \cr
  A\oplus B \oplus (A\odot B)
\end{array}
\right.
\]
For example, the first vector belongs to the characteristic $A\odot B$. It is a subset of $A$, of $B$, and of $A\oplus B \oplus AB$. All other sets do not include $A\odot B$. The ordering of the values of the entries in each basis vector of the kernel of a graph Laplacian accounts for the ``strength of uniqueness'' of the characteristic with regard to the given Boolean ring element.  By sorting the elements according to the eigenvectors of the graph Laplacian (for eigenvalue $0$) we figure out, to what extend certain characteristics are representative for the whole object: $A\odot B$ represents $A$ and $B$ by only 50\%, it represents $A\oplus B \oplus (A\odot B)$ by 33\%. 

\begin{figure}[h]
    \centering
    \includegraphics[width=0.5\linewidth]{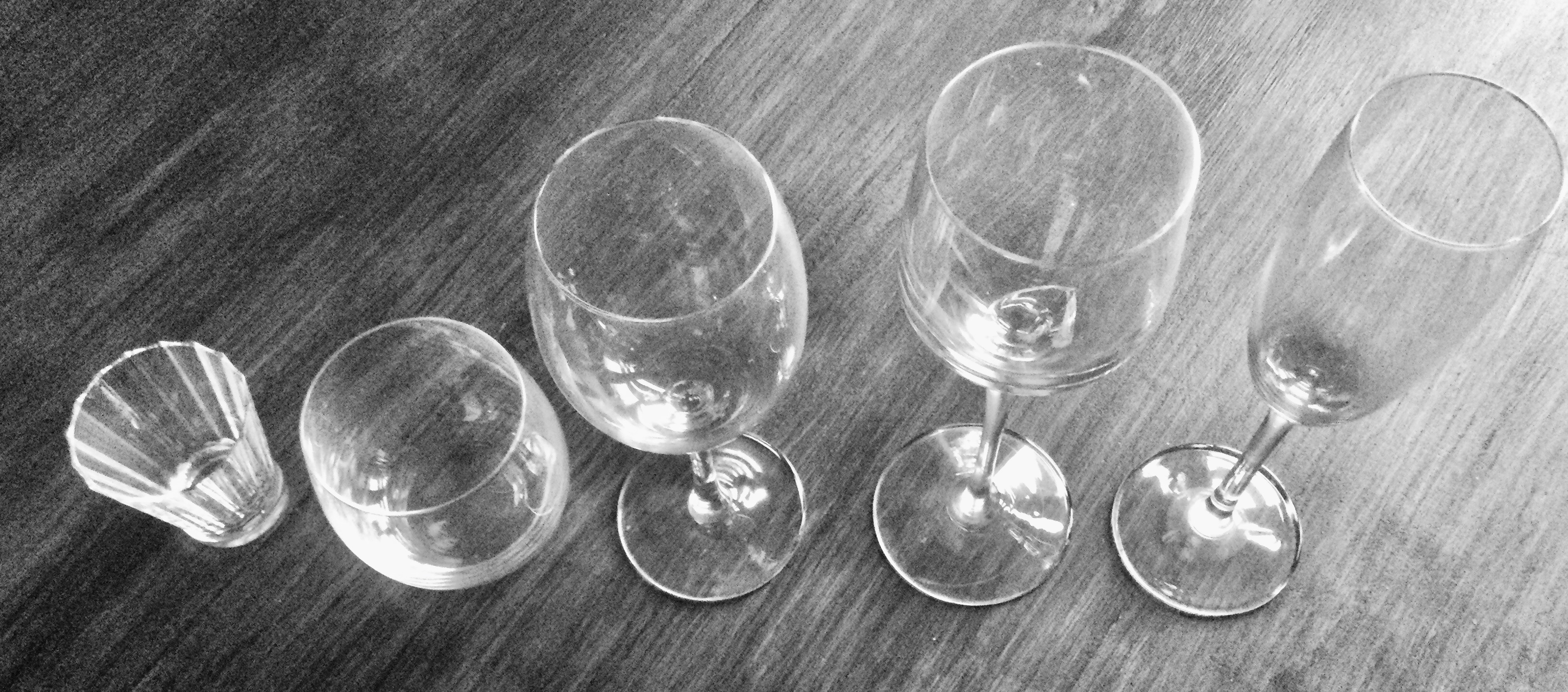}
    \caption{Search for the separation point in this sorted row! You will see, that almost ``naturally'' your brain tries to identify different binary characteristics (which define the shape of the glasses) when searching for delimitation.}
    \label{glasses}
\end{figure}

\paragraph{Sorting objects is a complexity reduction.} Given a set of $m$ objects, $i_1, \ldots, i_m$, the overall task is to find relevant characteristics. In this section, characteristics are assumed to be unavailable at the beginning of our analysis. The given task is then related to the task of grouping the objects into meaningful clusters. Once the clustering of objects is given, one can start to extract the differences and commonalities which are ``behind'' this clustering. Let us assume that we aim at a clustering into two groups, then we can find $2^{m-1}$ different possible clusterings. This number can be reduced: Suppose the objects are arranged in a sorted row, as is the case in Fig.~\ref{glasses}. Then clustering of the objects by finding a separation point in this sorted row reduces the clustering problem to only $m$ possible solutions. 

This type of complexity reduction is aimed at in the next scheme. The algorithm will evaluate formulas of the type $a\oplus (a\odot b)$ which are empty sets whenever $b\geq a$. The $\rho$-function turns these expressions into positive numbers which are high, whenever $b\geq a$ is ``valid'', and which are low, whenever $b\geq a$ is not ``valid''. Thus the resulting matrix $\bf R$ represents a numerical pendant of an adjacency matrix of the $\geq$-relation between the input objects. In contrast to $\bf L$ the matrix $\bf R$ is constructed such that it only has non-degenerate eigenvalues.  In this case, the number of ``clusters'' can be chosen and we choose $2$ clusters. The assumption of $2$ clusters means that there are two dominating discriminating characteristics (and we expect two committor functions).

\paragraph{Guiding example.} The realization of the scheme is illustrated with a simple example of the fairy tale ``Hansel and Gretel'' by the Brothers Grimm. The ``objects'' to be clustered will be given by the five protagonists of the story. The analysis is performed on the German version \cite{grimm} of this fairy tale to be found in internet, e.g. at \url{https://www.grimmstories.com/de/grimm_maerchen/hansel_und_gretel}. The objects are 

\vspace*{0.5cm}
$i_1$ = Hansel, $i_2$ = Witch, $i_3$ = Stepmother, $i_4$ = Gretel, $i_5$ = Father.
\vspace*{0.5cm}

\paragraph{The matrix {\bf M}.} The basis for clustering is a matrix $\bf M$. The element ${\bf M}_{jk}$ contains all commonalities between the $j$th and the $k$th object. Using the Boolean notation, the elements of the symmetric matrix $\bf M$ can be represented as ``products'' of two coded objects, namely: 
\begin{equation}
    {\bf M}_{ij}=f(i_j)\odot f(i_k)=y_j\odot y_k,
\end{equation}
where $y_k=f(i_k)$. For each pair of objects an expert has to determine their commonalities. This provides the following simplified results: 
\begin{itemize}
\item[] ${\bf M}_{12}=y_1\odot y_2 =$ ``instructions to Gretel; own plans are sabbotaged; is fooling others'' (Hansel and Witch)
\item[] ${\bf M}_{13}=y_1\odot y_3 =$ ``uses lies to carry out plans; pretends to keep others safe'' (Hansel and Stepmother)
\item[] ${\bf M}_{14}=y_1\odot y_4 =$ ``a child; adults seek for murder; escapes with cunning; gets rich in the end'' (Hansel and Gretel)
\item[] ${\bf M}_{15}=y_1\odot y_5 =$ ``worries about Gretel; male; instructions to Gretel; tries to escape a difficult situation; gets rich in the end'' (Hansel and Father) 
\item[] ${\bf M}_{23}=y_2\odot y_3 =$ ``does not have a particularly emotional bond with the children; seek the death of the children; uses cunning, pretends to be nice; dies in the end'' (Witch and Stepmother)
\item[] ${\bf M}_{24}=y_2\odot y_4 =$ ``a female role; homework; is capable of murder; uses 'pretending to be innocent' as a trick'' (Witch and Gretel)
\item[] ${\bf M}_{25}=y_2\odot y_5 =$ ``is able to let children die; adult person; gives instructions'' (Witch and Father)
\item[] ${\bf M}_{34}=y_3\odot y_4 =$ ``female role and attributes (homeliness); pretends to be understanding; is capable of murder'' (Stepmother and Gretel)
\item[] ${\bf M}_{35}=y_3\odot y_5 =$ ``plans to abandon the children in the forest; has power over the children; suffers from hunger'' (Stepmother and Father)
\item[] ${\bf M}_{45}=y_4\odot y_5 =$ ``gets rich in the end; has a closer relationship with Hansel; accepts death of others; has open uncertainties'' (Gretel and Father)
\end{itemize}
The thoughts about the similarities between the individual characters of the fairy tale arose spontaneously when the specific task was given that certain characters should be compared. Before this concrete ``duty'' to examine commonalities, we were not yet aware of some thoughts. For example, it was only in the course of the analysis that it became clear that Gretel is also capable of killing a person, while Hansel never has to make this specific decision. The textual results are not able to represent {\em all} possible thoughts. They can be regarded as mnemonics to be able to remember ``everything'' in the later steps of the algorithm.

\paragraph{The matrix $\bf R$.} The task of the second step (the $\rho$-part) is to turn the entries of the matrix $\bf M$ into numbers resulting in a positive real matrix $\bf R$. The entries of the matrix $\bf R$ are rating numbers. For Step 2, we iteratively take each object of our set $I$ into consideration. For every object we look at those descriptions written down in $\bf M$ in which this specific object has been compared to another object. Now it is rated, how well this description suits to the corresponding object. This evaluation does not take place solely on the basis of the specific mnemonics, but on the basis of the ``underlying'' thoughts. In principle, we make ourselves again aware of the thoughts that brought us to the formulation of the above mnemonics and check the relevance of these thoughts in relation to the object. The concrete rating is based on positive real numbers with a predefined upper bound. This upper bound forces us to (not expressly) ``justify'' ratings which are lower than this value. This means that the expression ${\bf M}_{jk}\oplus y_j= (y_j \odot y_k) \oplus y_j$ is determined when rating the entries of $\bf M$, if we determine what is missing in ${\bf M}_{jk}$ when describing $i_j$. This is, however, a subjective rating. The evaluation would be ``objective'' if the object could decide for itself how well the mentioned features characterize it. In the end these subjective ratings are inserted into the off-diagonal elements of a matrix $\bf R$.

The described procedure is performed with the entries of the matrix $\bf M$. The results are shown in Tab.~\ref{tabM}. For every character of the story we listed the corresponding products $y_j\odot y_1, \ldots, y_j\odot y_5$ (using that $y_j\odot y_k=y_k\odot y_j$). In the third column of this table, it has been rated (on a scale between $1$ and $10$), how well the thoughts $y_j \odot y_k$ (for $j\not = k$) fit to the object $i_j$.  There are some easy decisions for the ratings like ${\bf R}_{12}\leq {\bf R}_{15}$, because ``instructions  to  Gretel;  own  plans  are  sabbotaged;  is  fooling others'' is less informative than ``worries about Gretel; male; instructions to Gretel; tries to escape a difficult situation; gets rich in the end''.  
\begin{table}

    \begin{tabular}{|l|p{0.73\linewidth}|r|r|}
    \hline
    Hansel & instructions to Gretel; own plans are sabbotaged; is fooling others &  3 & ${\bf R}_{12}$\cr 
    \hline
     & uses lies to carry out plans; pretends to keep others safe & 3 & ${\bf R}_{13}$\cr
    \hline     
     & a child; adults seek for murder; escapes with cunning; gets rich in the end & 10 & ${\bf R}_{14}$\cr
    \hline
     & worries about Gretel; male; instructions to Gretel; tries to escape a difficult situation; gets rich in the end & 10 & ${\bf R}_{15}$\cr
    \hline
    \hline
    Witch & instructions to Gretel; own plans are sabbotaged; is fooling others & 3 & ${\bf R}_{21}$\cr
    \hline
     &does not have a particularly emotional bond with the children; seek the death of the children; uses cunning, pretends to be nice; dies in the end  & 10 & ${\bf R}_{23}$\cr
    \hline     
     & a female role; homework; is capable of murder; uses 'pretending to be innocent' as a trick & 5 & ${\bf R}_{24}$\cr
    \hline
     & is able to let children die; adult person; gives instructions & 3 & ${\bf R}_{25}$\cr
    \hline
    \hline
    Stepm. & uses lies to carry out plans; pretends to keep others safe & 3 & ${\bf R}_{31}$\cr
    \hline
     & does not have a particularly emotional bond with the children; seek the death of the children; uses cunning, pretends to be nice; dies in the end & 10& ${\bf R}_{32}$\cr
    \hline     
     & female role and attributes (homeliness); pretends to be understanding; is capable of murder & 3 & ${\bf R}_{34}$\cr
    \hline
     & plans to abandon the children in the forest; has power over the children; suffers from hunger & 5 & ${\bf R}_{35}$\cr
    \hline
    \hline
    Gretel & a child; adults seek for murder; escapes with cunning; gets rich in the end & 10 & ${\bf R}_{41}$\cr
    \hline
     &  a female role; homework; is capable of murder; uses 'pretending to be innocent' as a trick & 3 & ${\bf R}_{42}$\cr
    \hline     
     & female role and attributes (homeliness); pretends to be understanding; is capable of murder & 2 & ${\bf R}_{43}$\cr
    \hline
     & gets rich in the end; has a closer relationship with Hansel; accepts death of others; has open uncertainties & 3 & ${\bf R}_{45}$\cr
    \hline
    \hline
    Father & worries about Gretel; male; instructions to Gretel; tries to escape a difficult situation; gets rich in the end & 8 & ${\bf R}_{51}$\cr
    \hline
     &is able to let children die; adult person; gives instructions & 3 & ${\bf R}_{52}$\cr
    \hline     
     & plans to abandon the children in the forest; has power over the children; suffers from hunger & 8 & ${\bf R}_{53}$\cr
    \hline
     & gets rich in the end; has a closer relationship with Hansel; accepts death of others; has open uncertainties & 3 & ${\bf R}_{54}$ \cr
    \hline
    \end{tabular}
    \caption{Computation of the matrices $\bf M$ and $\bf R$.}
    \label{tabM}

\end{table}

These rating numbers are inserted into the corresponding off-diagonal elements of a matrix ${\bf R}$. The diagonal elements of $\bf R$ are adjusted such that the row sums of $\bf R$ are always identical (here $26$):
\begin{equation}
    {\bf R}=\begin{pmatrix}
     0 & 3 & 3 & 10 & 10\cr
     3 & 5 & 10 & 5 & 3 \cr
     3 & 10 & 5 & 3 & 5\cr
     10 & 3 & 2 & 8 & 3\cr
     8 & 3 & 8 & 3 & 4
    \end{pmatrix}.
\end{equation}
The $\bf R$-step of the algorithm offers a possibility of collaborations. Asking a group of experts to do a rating can add a statistical justification to the rating numbers. It is not a statistics with regard to many objects. It is a statistics with regard to experts opinions. 

\paragraph{Schur decomposition.} We employ the Schur decomposition of the matrix ${\bf R}={\bf U}{\bf V}$. 
Applying the Schur decomposition on $\bf R$ returns an orthogonal matrix $\bf U$ and an upper triangular matrix $\bf V$. According to the sorted Schur method \cite{Brandts} the eigenvalues on the diagnonal of $\bf V$ are arranged in an descending order. In our case they are sorted according to the absolute distance from the highest eigenvalue. Due to the theorem of Perron for positive matrices, the eigenvalue of $\bf R$ with largest absolute value is real and simple. Due to construction, the highest eigenvalue corresponds to the row sum of $\bf R$. Note that adding a multiple of the unit matrix to $\bf R$ does not change its eigenvectors. It also does not change the order of the eigenvalues of $\bf R$. This is the mathematical reason, why it is not important which specific row sum is chosen in {\bf R}. Also a negative entry on the diagonal would be possible, such that the row sums of $\bf R$ are zero. In this case, $\bf R$ can be seen as the graph Laplacian of a directed weighted graph. As it has been described in \cite{Brandts} a sorting of eigenvalues also has consequences for the corresponding Schur vectors in the matrix $\bf U$. From this matrix $\bf U$ the second column is taken into consideration. The objects are sorted according to the values of the corresponding vector $u_2$. More precisely, a vector $vec$ is computed by scaling and shifting the entries of $u_2$, such that the entries of $vec$ are in the range from $0$ to $1$. Furthermore, two other matrices are now computed. One two-columned matrix is $\chi$. The first column of this matrix is $vec$ and the second column is $1-vec$ (i.e., $\chi_{i2}=1-vec_i$), this matrix represents the ``committor functions''. The other matrix is the $2\times 2$ matrix ${\bf R}_c=(\chi^\top \chi)^{-1}\chi^\top {\bf R} \chi$.


\paragraph{Alternative mathematical justification of eigenvectors.} The non-symmetry of $\bf R$ stems from the non-symmetry of $\geq$-relations. In our example the matrix $\bf R$, however, is ``nearly'' symmetric. This could have a ``psychological'' reason. If we are asked to give ratings, then we already have a kind of (not explicitly formulated) numerical assessment scheme in mind, such that the occurrence of specific content in the description leads to a specific added contribution to the rating. For example: ``plans to abandon the children in the forest; has power over the children; suffers from hunger'' could be rated like: ``plans to abandon the children in the forest = 4 points''; ``has power over the children = 1 point''; ``suffers from hunger = 3 points''. Such a rating scheme would lead to a rating of $8$ for ${\bf R}_{53}$ as well as for ${\bf R}_{35}$. If the matrix $\bf R$ which is assumed to be an approximation of an adjacency matrix of a $\geq$-relation is {\em symmetric}, then the entries of $\bf R$ represent ``$=$'' (high numbers) or ``$\not=$'' (low numbers) relations. Thus, $\bf R$ is modelling a similarity matrix which also allows for a different interpretation of the role of the eigenvectors of $\bf R$ in terms of spectral clustering.   

There exists a list \cite{EuroVis}  of different methods which aim at reordering a matrix $\bf R$, such that the reordered matrix $\widetilde{\bf R}$ reveals a hidden block-structure in $\bf R$. Here, the hidden block structure of $\bf R$ (resp. $\bf M$) is identified by reordering its rows and columns according $vec$. The linear algebra used is similar to the ideas of GenPCCA, cf.~\cite{GenPCCA}. The intended order of objects is possible by sorting the entries of this vector $vec$.  
This is the way how the total order of objects is constructed. What is the reasoning behind this method?

Note that the routine also computes a matrix ${\bf R}_c$. It is the product of the pseudo-inverse of $\chi$, given by $(\chi^\top \chi)^{-1}\chi^\top$, multiplied with $\bf R$ and $\chi$. The matrix $\chi$ represents a fuzzy clustering of the objects. We will see, that the first column of $\chi$ represents a cluster {$A$} and the second column represents  a cluster {$B$}. The factor $(\chi^\top \chi)^{-1}$ normalizes ${\bf R}_c$. This normalization is such that the row sum of ${\bf R}_c$ equals the row sum of $\bf R$. This follows from the fact, that the row sums of $\chi$ are one and that the row sums of $\bf R$ are all equal.  

$\chi$ would be denoted as a ``crisp'' clustering, if all entries of $\chi$ would either be $0$ or $1$. In this case, the factor $(\chi^\top \chi)^{-1}$ would be a diagonal normalization matrix for the rows of ${\bf R}_c$. The matrix $\chi^\top {\bf R}\chi$ is then like summing up the ratings inside the clusters. The diagonal elements of ${\bf R}_c$ would then be like the mean of the rating numbers within the clusters. The off-diagonal elements would correspond to the mean rating numbers between the clusters.  Now, the actual $\chi$ is a relaxation of this strict $\{0,1\}$-assignment to the clusters by allowing for values in the interval $[0,1]$. It is a fuzzy clustering. Thus, ${\bf R}_c$ is like the projection of the rating matrix onto a $2\times 2$-rating matrix between the fuzzy clusters, cf.~\cite{GenPCCA1}. The task is to find a clustering $\chi$, such that ${\bf R}_c$ is as much as possible a diagonal matrix with minimal off-diagonal entries. One can, e.g., look for the matrix $\chi$ which maximizes the determinant or the trace of ${\bf R}_c$. The determinant and the trace of ${\bf R}_c$ are equal to the product, respectively sum, of the eigenvalues of ${\bf R}_c$. The presented method determines an invariant subspace of the matrix $\bf R$ spanned by the two leading eigenvectors of the matrix. A basis of this invariant subspace is given by the two column vectors in the matrix $\chi$. This is due to the fact, that the columns of $\chi$ are a scaled and shifted version of the second Schur vector. Thus, it is also a linear combination of the first (constant) and the second eigenvector of $\bf R$. By this construction, the matrix ${\bf R}_c$ inherits the highest eigenvalues from $\bf R$.  The aim of the presented method is to provide an assignment of the objects to the clusters (in a fuzzy sense), such that the matrix ${\bf R}_c$ is as close as possible to a diagonal matrix. The described procedure is a spectral clustering method, cf.~\cite{UL}.  

The problem of finding clusters in {\bf R} turns into a graph partitioning problem, cf.~\cite{GraphPart}. The problem of finding the optimal clustering (minimal cut) of this graph is, e.g., solved by looking at the eigenvector of the second largest eigenvalue of this matrix and by separating the vertices with negative entry from the vertices with positive entry (Fiedler's cut, cf.~\cite{Fiedler}). Note, that the presented method is equal to computing the second largest eigenvector and rescaling and shifting its entries. Thus, the order of elements stays the same like in Fiedler's cut.  In this section it has been assumed that the second highest eigenvalue is a real number. If there is a second ``largest'' conjugate pair of complex eigenvalues, then it is questionable, whether the assumption of having two fuzzy clusters is valid. Since in this case, the matrix $\bf R$ could be seen as a multiple of a transition matrix of a non-reversible Markov chain, which would allow for a $3$-clustering via GenPCCA \cite{GenPCCA}\cite{GenPCCA1}. 

\paragraph{Results.} Applying the Schur decomposition to the matrix $\bf R$ leads to the following shifted and rescaled output: 
\begin{equation}\label{eq:vec}
    vec=\begin{pmatrix}0.67003 \cr
0.10815 \cr
0.00000 \cr
1.00000 \cr
0.27644
\end{pmatrix},\quad 
{\bf R}_c=\begin{pmatrix}   15.8541  & 10.1459\cr
    7.4738 &  18.5262 \end{pmatrix}.
\end{equation}
On the one hand, one can already derive the sorted row of objects from the vector $vec$ by just sorting its entries, namely:

\vspace*{0.5cm}
Gretel, Hansel, Father, Witch, Stepmother.
\vspace*{0.5cm}

On the other hand, the matrix ${\bf R}_c$ indicates that the coherence of one cluster (where Gretel is part of) is lower than the coherence of the other cluster (where the Stepmother is part of). This is, because the first diagonal element of ${\bf R}_c$ is smaller than the second diagonal element. In general, small off-diagonal elements in ${\bf R}_c$ indicate a good separation of the clusters. 
The ``most substantial cut'' in this sorted row of objects (Gretel, Hansel, Father, Witch, Stepmother) is maybe between Father and Witch. Gretel, Hansel, and Father get rich in the end, whereas the Witch and Stepmother do not have a particularly emotional bond with the children and they die in the end. The property ``emotional bond with the children'' is a kind of non-binary characteristic, because it cannot be answered with ``yes'' or ``no'' for the children. However, this property characterizes the Witch and the Stepmother and maybe is the ``moral'' reason why they have to die in the end, while the ``real'' family gets rich. Once the clustering Gretel, Hansel, Father versus Witch and Stepmother is given, characteristics (``real family'') quickly come to mind that were previously only latently echoed in the mnemonics and had not been written down yet. And suddenly it also becomes clear why there is that part of the fairy tale in which Gretel asks her brother to only use the duck one at a time to swim across the lake. The three people who ultimately showed compassion will be rewarded for it in the end.

\paragraph{Alternative interpretation of $u_2$.} The matrix $\bf R$ is non-symmetric. The weighted summing up of the elements of $\bf R$ with respect to the two clusters by means of $\chi^\top {\bf R}\chi$ does, in general, also not lead to a symmetric matrix.  What kind of changes $\bf D$ adjust the ratings in $\bf R$ such that its row sums are still identical and such that the matrix $\chi^\top ({\bf R} + {\bf D})\chi$ (i.e., the summed rating on the level of the clusters) is symmetric, while keeping the clustering $\chi$ fixed? Necessarily, to every row of $\bf R$ we have to add a row vector such that the sum of the elements of this vector is zero. Is there a possibility to add the {\em same} vector to {\em every} row? The following arguments are based on the assumption that the second largest eigenvalue of $\bf R$ is real. We will see that in principle the ``Fiedler vector'' is doing this job. In our case it is the second Schur vector, i.e. the second column of $\bf U$ denoted as $u_2$. The first Schur vector $u_1$ is a constant vector. The $i$-th column of $\chi$ can thus be written as $\alpha_i u_1+\beta_i u_2$ with specific real numbers $\alpha_i$ and $\beta_i$. Let $v_{12}$ denote the first off-diagonal element of the matrix $\bf V$ in the corresponding Schur decomposition ${\bf R}={\bf U}{\bf V}$. The dyadic product ${\bf D}=-v_{12} u_1u_2^\top$ is a matrix which has identical rows and which has row sum zero, because $u_1$ is constant vector and it is orthogonal to $u_2$. To check the symmetry of ${\bf C}=\chi^\top ({\bf R}+{\bf D})\chi$, we compute its elements:
\begin{eqnarray}
\nonumber
(\chi^\top ({\bf R}+{\bf D}) \chi)_{ij} & = & (\alpha_i u_1 +\beta_i u_2)^\top ({\bf R}-v_{12} u_1 u_2^\top)(\alpha_j u_1+\beta_j u_2)\cr
& = & (\alpha_i u_1 +\beta_i u_2)^\top ({\bf R}(\alpha_j u_1+\beta_j u_2)-\beta_j v_{12} u_1)\cr
& = & (\alpha_i u_1 +\beta_i u_2)^\top (v_{11}\alpha_j u_1+v_{22}\beta_j u_2+\beta_j v_{12} u_1-\beta_j v_{12} u_1)\cr
& = & (\alpha_i u_1 +\beta_i u_2)^\top (v_{11}\alpha_j u_1+v_{22}\beta_j u_2)\cr
& = & v_{11}\alpha_i \alpha_j +v_{22}\beta_i\beta_j.
\end{eqnarray}
Thus, $\bf C$ is symmetric. The elements of $u_2$ denote an {\em object-based} adjustment of the ratings such that the matrix $\bf C$ is symmetric. The matrix ${\bf R}+{\bf D}$ has the same invariant subspace (with the same eigenvalues) like $\bf R$. $u_2$ is even an eigenvector of ${\bf R}+{\bf D}$. The ordering of the objects according to $vec$, therefore, also takes the non-symmetry of ${\bf R}$ into account. It takes into account, that the ratings in $\bf R$ are not just based on weighted sums of assessed characteristics and that ratings in principle are based on (partial) $\geq$-relations.

\subsection{Feature extraction}\label{sec:fs}
In the next example, we already have a list of binary characteristics, such that we can answer for every object whether a certain characteristic applies or not. The objects are represented by strings of bits via $f:I\rightarrow \cal B$. Every bit of the string represents a possible characteristic. For a given object, this characteristic either applies (1) or not (0).  

\paragraph{Linear regression.} For constructing the function $\rho:{\cal B}\rightarrow O$, every bit of the string $y\in \cal B$ has a (yet unknown and to be learned) weight. The intended linear function simply sums up all weights which belong to the $1$-bits of $y$.  We adjust the weights such that the output value of $\rho$ is as close as possible to a given intended value (in a mean-square-distance manner with non-negativity constraints for the weights). In order to provide a simple example for this approach, we take the five different characters of the fairy tale into account again: $i_1$ = Hansel, $i_2$ = Witch, $i_3$ = Stepmother, $i_4$ = Gretel, $i_5$ = Father. 
For the training phase we need to provide the intended values of $\rho$. In our case, we want to understand what creates the coordinates Hansel$=0.67003$, Witch$=0.10815$, Stepmother$=0.00000$, Gretel$=1.00000$, Father$=0.27644$ from the previous section, Eq.~(\ref{eq:vec}). 

\paragraph{Number of equations.} At this point, everything looks very straight forward. Every object $i_k\in I$ has a certain intended value $g(i_k)=\rho(f(i_k))\in O$. The weights should quantify the importance of certain characteristics. This situation might be seen as the starting point of a machine learning algorithm. In the case of linear regression, the number of linear equations to be formulated is equal to the number of objects, whereas the number of unknowns equals the number of binary characteristics. This means that we will probably end up with more unknowns than equations if we analyze complex objects. The rank of the coefficient matrix will probably be lower than the number of unknowns. 

In the beginning we started with the fundamental idea that the importance of characteristics is only visible by taking comparisons of objects into account. 
At this stage, the Boolean operations $\oplus$ and $\odot$ come into play (in between $f$ and $\rho$). They correspond to bit-wise applications of XOR and AND to the strings of bits. They can be performed by computers and do not need extra expert knowledge. Let us assume that we want to quantify the similarity of the objects, i.e., we want to apply AND to the bits of the strings. The intended output values of $\rho$ can be chosen according to a predefined (expert) model for quantifying similarity. The similarity value of two input objects $i_a$ and $i_b$ will be modeled as $$\rho(f(i_a)\odot f(i_b))=e^{-10\cdot \|x_a-x_b\|^2},$$ where $x_a$ and $x_b$ are the corresponding coordinates of the objects. Instead of having $n$ equations for intended values, we now have an order of $n^2$ equations. This might be a good reason for using the fundamental scheme in Fig.~\ref{fig:origmap} also for machine learning algorithms in order to extend the number of training points, whenever we deal with very complex objects. Not the $n$ objects themselves are in the training set, but the modelled $O(n^2)$ values of Boolean expressions.  

\paragraph{Example.} For illustrating the new linear regression approach, the given binary characteristics should be: ``female'', ``adult'', ``dies in the end'' and ``part of the family''.  The interesting point is, that this list of characteristics can be extended by taking also the opposite characteristics into account: ``male'', ``non-adult'', ``survives in the end'' and ``non-family''.  The $5$ strings of bits are:
$y_1=f(i_1)=00011110$, $y_2=f(i_2)=11100001$, $y_3=f(i_3)=11110000$, $y_4=f(i_4)=10010110$, and $y_5=f(i_5)=01011010$. For every bit of the string a weight has to be learned -- in total $8$ weights $w_1,\ldots w_8$. Table.~\ref{tab:my_label4} summarizes the training data.

\begin{table}[h]
    \vspace*{0.2cm}
    \centering
    \begin{tabular}{|l|c|c||c|}
         \hline
         sum of weights & intended $\rho$ &  expression & trained $\rho$ \cr
         \hline
          & 0.04254 &  $\rho(y_1\odot y_2)$ & 0.00000\cr
          $w_4$& 0.01122 &  $\rho(y_1\odot y_3)$& 0.00000\cr
          $w_4+w_6+w_7$& 0.33662 &  $\rho(y_1\odot y_4)$& 0.33662\cr
          $w_4+w_5+w_7$& 0.21244 &  $\rho(y_1\odot y_5)$& 0.21244\cr
          $w_1+w_2+w_3$& 0.88962 &  $\rho(y_2\odot y_3)$& 0.88962\cr
          $w_1$& 0.00036 &  $\rho(y_2\odot y_4)$& 0.00020\cr
          $w_2$& 0.75336 &  $\rho(y_2\odot y_5)$& 0.60954\cr
          $w_1+w_4$& 0.00004 &  $\rho(y_3\odot y_4)$& 0.00020\cr
          $w_2+w_4$& 0.46572 &  $\rho(y_3\odot y_5)$& 0.60954\cr
          $w_4+w_7$& 0.00532 &  $\rho(y_4\odot y_5)$& 0.00532\cr
         \hline
    \end{tabular}
    \vspace*{0.2cm}
    \caption{The training data for learning sums of weights $w_1,\ldots,w_8$ (first column) on the basis of the given intended similarity values (second column) for the commonalities of the characters of the fairy tale (third column). After having learned the weights the actually trained sums are shown in the last column.}
    \label{tab:my_label4}
\end{table}
 
\paragraph{Results.} The learned weights are $w_1= 0.00020, w_2=0.60954, w_3=0.27988,$ $w_4=0.00000, w_5=0.20711, w_6=0.33130, w_7=0.00532$. Note, that $w_8$ need not be learned, because it does not occur in the equations. The rank of the coefficient matrix for $w_1,\ldots, w_7$ is full and equals the number of unknowns.   

The most important commonality (given by $w_2$ ad $w_6$) of two figures of the fairy tale according to the proposed coordination is ``being adult/non-adult''. This characteristic determines at most, whether the coordinate values of the persons are nearby or not.  This most important characteristic is followed by $w_3$ for ``dying in the end''. Regarding the sorted row (Gretel, Hansel, Father, Witch, Stepmother) these characteristics really correspond to the grouping principles. The same holds for the next important characteristic ``male'': The males (Hansel and Father) are grouped together in this row, but have already rather different ``coordinate values''. In contrast to ``male'', the characteristic ``female'' is not important for the grouping of objects, it also has a low weight $w_1=0.00020$.

\subsection{Algebraic feature extraction}\label{sec:afs}

In statistical methods, a rule about objects is assumed to be valid, if a certain relation or pattern is observed often. The philosophy of algebraic feature selection is, that all imaginable relations between the object properties apply, unless one finds an object which contradicts such a relation. From this point of view, observing many objects does not generate relations. In fact, every new observed object destroys possible relations. In accordance with statistical methods, the algebraic feature selection works if one has collected enough data that comprehensively represent all possible relations. 

\paragraph{Formal Concept Analysis.} Formal Concept Analysis (FCA) \cite{Rille} is also a realization the fundamental scheme in Fig.~\ref{fig:origmap}. The input objects are already coded. Thus, like in Sec.~\ref{sec:fs}, the results of the function $f$ are already available. On the basis of the given elements $f(i_j)$ of a Boolean ring, FCA (in principle) uses an algorithm to compute an output element $z$ of the Boolean ring which serves as a generator of an ideal $\langle z \rangle$. FCA represents this element $z$ in form of a diagram, which is the final mapping $\rho(z)$. In this section, we will work out the relation between FCA and algorithms in Boolean rings.     

FCA is based on tables in which the assignment of categories is collected, like in Table \ref{tab:FCA}. From this table a hierarchy of the used terms can be derived. The result of FCA is often presented by a diagram. An example for such a diagram is shown later.

\begin{table}[h]
    \centering
    \begin{tabular}{|c||c|c|c|c|c|}
              \hline
             & visible & audible & static & haptic & only ``techn. produced'' \\
              \hline
         book & 1 & 0 & 1 & 1 & 1\\
         video & 1 & 1 & 0 & 0 & 1\\
         picture & 1 & 0 & 1 & 1 & 1\\
         sound & {\bf 0} & 1 & 0 & 0 & 0\\
         \hline
    \end{tabular}
    \caption{Properties of different media. Tables of this kind are the starting point of a formal concept analysis.}
    \label{tab:FCA}
\end{table}

First of all, two observations can be made. On the one hand, the structure of Table~\ref{tab:FCA} shows that each row (and also each column) is a list of zeros and ones. Thus, they can also be understood as elements of  Boolean rings. Second, it becomes clear that in FCA there is a distinction between terms denoting objects and terms denoting features. The sentence ``The book is visible'' structures our thinking. ``Book'' becomes a term which seems to denote an object and ``visible'' becomes a term which denotes its characteristic. However this distinction is only due to our special grammatical form used to formulate this sentence. ``Being a book'' and ``being visible'' are actually both just terms that are used to differentiate the specific object from other objects, or to express similarities with other objects (in terms of usability and visibility). If you were to demonstrate two visible objects and say ``This visible is book and this visible is towel'' then it becomes clear that ``object names'' are actually object properties as well. The structure of FCA suggests that for terms denoting objects, we need to find the corresponding object properties that are common to all objects sharing the same ``name'' (e.g. ``books'').  However, we want to start here from the situation in which there are concrete objects for which relevant distinguishing criteria are sought, with the object names being part of the object properties. Taking all possible object properties into account and checking for each object whether they apply nor not, we arrive at the same situation like in FCA, i.e., each object is represented by an element of a Boolean ring (a list of zeros and ones). 

{\tiny
\begin{table}[h]
    \centering
    \begin{tabular}{|c|ccccccc|}
              \hline
             & book & video &  sound & visible & audible & red & produced \\
              \hline
         object1 & 1 & 0 &  0 & 1 & 0 & 1 & 1\\
         object2 & 0 & 1 &  0 & 1 & 1 & 0 & 1\\
         object3 & 0 & 0 &  1 & 0 & 1 & 0 & 1\\
         \hline
    \end{tabular}
    \caption{Different concrete objects are collected and properties of these objects are assigned to them.}
    \label{tab:Boolean}
\end{table}
}

As a guiding example, we will take a more artificial table (Table \ref{tab:basis}).

\begin{table}[h]
    \centering
    \begin{tabular}{|c||c|c|c|c|}
              \hline
              & A & B & C & D\\
              \hline
              \hline
         object1 & 1 & 0 &  0 & 0 \\
         \hline
         object2 & 0 & 1 &  1 & 0 \\
         \hline
         object3 & 0 & 1 &  0 & 1 \\
         \hline
         object4 & 0 & 1 &  0 & 0 \\
         \hline
    \end{tabular}
    \caption{This table is the guiding example for the further analysis. It is constructed in the same way like Table~\ref{tab:Boolean}.}
    \label{tab:basis}
\end{table}

\begin{figure}[h]
    \centering
    \includegraphics[width=0.4\textwidth]{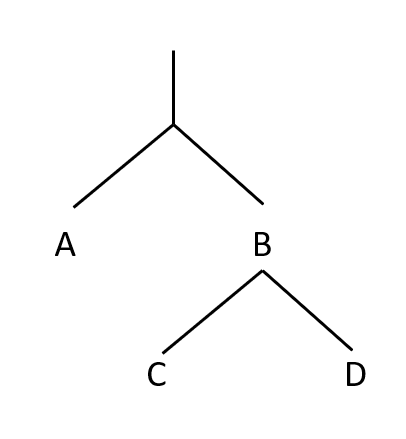}
    \caption{FCA diagram of the hierarchy of the terms A, B, C, and D in Table~\ref{tab:basis}.}
    \label{fig:my_label}
\end{figure}

\paragraph{Knowing the hierarchy.} Let us assume that the hierarchy of the terms (A, B, C, and D) used in Table~\ref{tab:basis} is already known and presented by the diagram in Figure~\ref{fig:my_label}.  

This diagram will now be used to reduce the information that is presented in the rows of Table~\ref{tab:basis}. The diagram can be transformed into an ideal of the corresponding Boolean ring consisting of the variables $A$, $B$, $C$, and $D$. The equations which can be derived directly from this diagram are: $AB=0$, because an object can not be $A$ as well as $B$. The same holds for the equation $CD=0$. The hierarchical aspects can be expressed using the subset properties $BC\oplus C=0$ and $BD \oplus D=0$. The last row of Table~\ref{tab:basis} shows, that it is possible for an object to have property $B$ without having property $C$ or $D$. If ``having $B$'' would automatically mean to be either ``C'' or ``D'', then a further equation would hold: $B=C\oplus D$ or equivalently $B\oplus C\oplus D=0$. 
The diagram in Figure~\ref{fig:my_label}, thus, defines equations (vice versa, equations also define diagrams of that type). The set of all equations leads to an ideal in the corresponding Boolean ring. Finding the generator of this principle ideal is possible by taking the ``union'' of all defining equations. In this case it leads to the ideal
\[
{\cal I}=\langle ABCD\oplus AB \oplus BC \oplus BD \oplus CD \oplus C \oplus D\rangle
\]
representing Figure~\ref{fig:my_label}, i.e., representing the hierarchy of terms. 

A reduction of information contained in a row of Table~\ref{tab:basis} is now performed by a division with remainder according to the ideal $\cal I$. Take for example object1. This object is defined by the equation
\begin{eqnarray}\label{eq:object1}
object1&=& A(B\oplus 1)(C\oplus 1)(D\oplus 1)\cr
&=& ABCD \oplus ABC \oplus ABD \oplus AB \oplus ACD \oplus AC \oplus AD \oplus A. 
\end{eqnarray}
Taking the remainder with regard to the ideal $\cal I$ leads to 
\[
object1 \equiv A \quad \mathrm{mod}\quad {\cal I}, 
\]
which means that the only relevant information about object1 is property ``A'' (further information about $B$, $C$, or $D$ is redundant). 

\paragraph{Missing entries.} It has been described, how a hierarchy of terms can be transformed into an ideal. The reduction of information about objects can be performed by determining the residue of corresponding products (\ref{eq:object1}) with regard to the ideal. Note that missing information (no assignment with ``0'' or ``1'' is possible) about object1 may be modelled by leaving out certain linear factors of the product in (\ref{eq:object1}).

\paragraph{Deriving the hierarchy.} Finding a hierarchy of terms is the result of many observations and the act of comparing objects. The reason for the fact that the columns ``sound'' and ``visible'' in Table~\ref{tab:Boolean} exclude each other is contained in the bold ``0'' in Table~\ref{tab:FCA} or in the fact that there simply does not exist an object which has the property ``sound'' as well as the property ``visible'' (the object would be a ``video'' in that case, and not a ``sound'').  

More philosophically speaking: the relations between properties are derived from missing objects. The fact that books are visible stem from our observation, that invisible books do not exist. The fact that we can imply that ``the ground becomes wet'' from the observation that ``it is raining'' is simply given by the missing situation (object) of ``a dry ground although it rains onto it''. The ideal $\cal I$ can be derived from our pre-knowledge (or rational) about the four properties in Figure \ref{fig:my_label}, however, it can also be derived from the missing objects in Table~\ref{tab:basis} without including pre-knowledge. The objects are defined by equations, see (\ref{eq:object1}). These equations can be seen as result of the mapping $f$ applied to the objects:
\begin{eqnarray*}
object2&=& (A\oplus 1)BC(D\oplus 1)\cr
&=& ABCD \oplus ABC \oplus BCD \oplus BC, \cr &&\cr
object3&=& (A\oplus 1)B(C\oplus 1)D\cr
&=& ABCD \oplus ABD \oplus BCD \oplus BD, \cr &&\cr
object4&=& (A\oplus 1)B(C\oplus 1)(D\oplus 1)\cr
&=& ABCD \oplus ABC \oplus ABD \oplus AB \oplus BCD \oplus BC \oplus BD \oplus B. 
\end{eqnarray*}
Now the algorithm is performed: The ``set'' of all existing observations is given by the union of the corresponding four equations defining the objects. This union is
\[
x = ABC\oplus ABD \oplus ACD \oplus AC \oplus AD \oplus A \oplus BCD \oplus B.
\]
The complement of this expression includes everything that has not been observed (so far). If we use that the union of $A$, $B$, $C$, and $D$
 is our ``unity'', and that this union is given by 
\begin{eqnarray*}
1 &= & ABCD \oplus ABC \oplus ABD \oplus AB \oplus ACD  \cr
  && \oplus AC \oplus AD \oplus A \oplus BCD \oplus BC \cr
  && \oplus BD \oplus B \oplus CD \oplus C \oplus D,
\end{eqnarray*}
then everything that has not been observed is $1\oplus x$, i.e.:
\[
1\oplus x= ABCD \oplus AB \oplus BC \oplus BD \oplus CD \oplus C\oplus D.
\]
This expression is exactly the generator of the ideal $\cal I$ representing the relationships between the objects $A$, $B$, $C$, and $D$. This generator has been derived from the missing observations and not from other sources of pre-knowledge. Figure~\ref{fig:my_label} can be seen as the output of $\rho(ABCD\oplus AB \oplus BC \oplus BD \oplus CD \oplus C \oplus D)$.

\paragraph{Proposed method of reducing information.} Given a table like Table~\ref{tab:basis}, the relevant information about the objects can be found in the following way. First, the algebraic expressions have to be generated from taking each object into consideration, like in  (\ref{eq:object1}). The union $x$ of these expressions has to be calculated. The ideal $\cal I$ is generated by the complement of this expression ${\cal I}=\langle 1\oplus x\rangle$. In order to find the relevant information about an object one has to determine ``its'' residue with regard to the ideal $\cal I$.  In the example above this leads to:
\begin{eqnarray*}
object1 &\mapsto& A\cr
object2 &\mapsto& C\cr
object3 &\mapsto& D\cr
object4 &\mapsto& B\oplus C \oplus D.
\end{eqnarray*}
This can be seen as an algebraic feature extraction method.

\paragraph{An example.} 
The scenario of the following example is from a study of ancient Egyptian objects that were found together early twentieth century in Karnak (a village close to Luxor) and which have been posing puzzles to experts for many years\footnote[1]{The full, large data set has more than 500 objects and more than 20 properties. It has been collected from an online database and expanded by cooperation partners from FU Berlin, Ralph Birk and Sarah Klasse}. Various teams of experts have already dealt with these objects. Each team has examined different aspects of the corresponding objects. 

In this way, a common large table is compiled. A tiny snippet  is given in Table.~\ref{tab:cachette} (it represents an evaluation of function $f$). 

\begin{table}[]
    \centering
    \begin{tabular}{|c|cc|cc|}
         
         \hline
         & \multicolumn{2}{c|}{France}&\multicolumn{2}{c|}{Germany}\\
         &  prop. (a) & prop. (b) & prop. (c) & prop. (d)\\
         \hline
         object 1 & 1 & 0 & 0 & 0 \\
         object 2 & 1 & 1 & 0 & 1 \\
         object 3 & 0 & 0 & 1 & 0 \\
         object 4 & 1 & 1 &   & 0 \\
         object 5 & 1 & 1 & 0 & 0 \\
         \hline
    \end{tabular}
    \caption{This table classifies found objects according to their types (prop. a and b) and the way they have been treated by the ancient people (prop. c and d).}
    \label{tab:cachette}
\end{table}

While one team of experts contributes the first two columns in Table~\ref{tab:cachette} (that is the team from France), the team from Germany  has added the last two columns. In the case of object 4, it is not clear whether the property c applies; the appropriate entry is therefore missing. The French team is not able to evaluate property c and d for yet unknown objects, whereas the German team would not be able to assign properties a and b correctly, because the experts of the teams stem from different disciplines. 

One task of collaboration is, for example, to gain knowledge by comparing objects and by finding delimitations between them. To do this, it is necessary to find similar objects (additional to the given ones) and include those in the investigation. But what would be comparable objects to object 1, for example? Which features characterize object 1? Thus, we use the algebraic feature selection method. The following applies due to Table~\ref{tab:cachette}:
\begin{eqnarray*}
object1 &=& a(b\oplus 1)(c\oplus 1)(d\oplus 1)\cr
object2 &=& ab(c\oplus 1)d\cr
object3 &=& (a\oplus 1)(b\oplus 1)c(d\oplus 1)\cr
object4 &=& ab(d\oplus 1)\cr
object5 &=& ab(c\oplus 1)(d\oplus 1).
\end{eqnarray*}

In this way (except for object 4) all objects are represented by atoms of the respective Boolean ring. The union of all these arithmetic expressions gives 
$$x = a\oplus abcd\oplus abd\oplus ad\oplus bc\oplus bcd \oplus c \oplus cd.$$
All relations that can be formulated on the basis of the data are part of the ideal ${\cal I}=\langle 1\oplus x\rangle$. For example, $cd$ is an element of $\cal I$. The equation $cd=0$ applies (there is no object with property c and also property d). Providing all relations between the properties would mean to provide all elements of $\cal I$. Up to this point everything is just based on the entries of the table, knowledge from ``outside'' is not included. 

The important features that characterize the objects are gained by division with remainder with respect to that ideal $\cal I$. The following applies:
\begin{eqnarray*}
object1 &\mapsto& a\oplus ab\cr
object2 &\mapsto& d\cr
object3 &\mapsto& c\oplus abc\cr
object4 &\mapsto& d\oplus ab\cr
object5 &\mapsto& ab\oplus abd \oplus abc.
\end{eqnarray*}

For objects which are also atoms of the Boolean ring the following interpretation of the algebraic features is possible: The algebraic feature selection provides a relation which would additionally hold (be part of $\cal I$), if that object would be removed from the list. If all objects are removed which have the same remainder like object 1, then setting $a$ would be the same as setting $a$ and $b$, i.e., $a\oplus ab=0$. This interpretation is possible, because an atom either ``completely belongs to the ideal $\cal I$'' or it  completely does not.  Removing object 2 would mean that the rule $d=0$ is added to the ideal $\cal I$. 
However, the representations for the above remainders are not unique. The freedom to transform the remainder needs an ``outside knowledge'' about what is understood to be a  ``convenient form''. This is something that is not given by the table. 

There is also an alternative interpretation of $x$. Multiplication with $x$ transforms object selections into an equivalent selection term. Multiplying $cd$ with $x$ provides zero. It is equivalent to say, that all objects are selected which share properties $c$ and $d$ in common, or to select the empty set. All selection rules $s$ which provide the atom $a(b\oplus 1)(c\oplus 1)(d\oplus 1)$ when multiplied with $x$ are valid remainders of object1. One example is $s=a\oplus ab$.

Now we can start looking for similar objects. ``Convenient'' algebraic transformations of the remainder even show how the work can be divided between the two teams, since certain properties can only be evaluated by the respective teams. The team from Germany would not be needed for selecting similar objects with regard to object 1, and the team from France would not be needed for object 2. For object 3 we have $c(ab\oplus 1)$, the German team would have to find objects that have the property $c$ and the French team would have to sort out those objects that have both $a$ and $b$. For object 5, one could write the equation as $ab(1\oplus c\oplus d$) and also split the work of finding similar objects accordingly. In this case the French team selects objects which have a and b, whereas the German team takes care that $d$ or $c$ are missing. These ``working plans'' represent the output $\rho$ of the fundamental scheme. The method is based on the assumption, that the table includes all important relations.

If this is not the case, then the ideal can also be used (in a different way). Then there are missing objects or missing relations: In order to find ``missing objects'', which would sort out wrong relations, one would have to find objects (outside the table) which have combinations of properties included in $\cal I$. $\cal I$ can thus be understood as a plan to find missing objects. On the other hand one could check, whether all known or presumed relations are to be found in $\cal I$. One could check, whether really all $b$-objects are also $a$-objects, i.e. whether $b(a\oplus 1)$ is an element of $\cal I$. The ``presumed relations'' are also outside the table. 

\paragraph{The dual concept.} Finally, we want to demonstrate that the philosophy of this manuscript (using the comparison of objects in order to formulate their properties) is a dual concept with regard to FCA. For this purpose we determine the transposed of Table~\ref{tab:basis} and arrive at Table~\ref{tab:dualbasis}. 

\begin{table}[ht]
    \centering
    \begin{tabular}{|c||c|c|c|c|}
              \hline
              & a & b & c & d\\
              \hline
              \hline
         property1 & 1 & 0 &  0 & 0 \\
         \hline
         property2 & 0 & 1 &  1 & 1 \\
         \hline
         property3 & 0 & 1 &  0 & 0 \\
         \hline
         property4 & 0 & 0 &  1 & 0 \\
         \hline
    \end{tabular}
    \caption{This is the transposed of Table~\ref{tab:basis}. The roles of objects and properties are exchanged.}
    \label{tab:dualbasis}
\end{table}
The algebraic approach is now the same as above, however, we exchanged the roles of objects and properties. The result will be an ideal, which can be used to reduce the effort of object comparisons. The described procedure can be seen as a method of how to formally derive the ideal in Sec.~\ref{sec:algo}. Using linear factors, the properties can be calculated as:
\begin{eqnarray*}
property1 &=& a(b\oplus 1)(c \oplus 1)(d\oplus 1)\cr
&=& abcd \oplus abc \oplus abd \oplus acd \oplus ab \oplus ac \oplus ad \oplus a, \cr &&\cr
property2&=& (a\oplus 1)bcd\cr
&=& abcd \oplus bcd, \cr &&\cr
property3&=& (a\oplus 1)b(c\oplus 1)(d\oplus 1)\cr
&=& abcd \oplus abc \oplus abd \oplus ab \oplus bcd \oplus bc \oplus bd \oplus b, \cr &&\cr property4&=& (a\oplus 1)(b\oplus 1)c(d\oplus 1)\cr
&=& abcd \oplus abc \oplus bcd \oplus acd \oplus bc \oplus ac \oplus cd \oplus c.
\end{eqnarray*}
There may be relations between the objects which can be used to simplify these ``computations'' of the characteristics. The union of the above expressions is
\[
 y = abc \oplus ad \oplus a \oplus bcd \oplus bd \oplus b \oplus cd \oplus c. 
\]
These are the ``observed properties''. The relation between the objects is given by the non-observed properties, which leads to the ideal:
\[
{\cal J}=\langle1\oplus y\rangle.
\]
The task is now to compute the remainder of the properties with regard to the ideal $\cal J$. In our example this leads to:
\begin{eqnarray*}
 property1 &\equiv& a \quad \mathrm{mod} \quad {\cal J}, \cr
 property2 &\equiv& bcd \quad  \mathrm{mod} \quad {\cal J}, \cr
 property3 &\equiv& b \oplus bcd \quad  \mathrm{mod} \quad {\cal J}, \cr
 property4 &\equiv& c \oplus bcd \quad  \mathrm{mod} \quad {\cal J}.
\end{eqnarray*}
Like in Sec.~\ref{sec:algo} we arrive at a method to define characteristics on the basis of a comparison of objects. In contrast to Sec.~\ref{sec:algo} the pre-knowledge about object relations is derived from existing annotation data. ``Property 4'', e.g., becomes visible by analyzing the difference between object $c$ and all commonalities of objects $b, c,$ and $d$. A delimitation from object $a$ is not needed. 

\section{Conclusion}
The possibility of mathematization exists in research fields that use object comparisons as a means of gaining knowledge - also exactly where gaining knowledge ``happens''. 

If we limit ourselves to ``number mathematics'', then we keep disciplines separate: The objects have to be coordinated by numbers, so we leave the original discipline and enter the field of arithmetic. After transforming the coordinates by algorithms, we deliver (or visualize) a numerical result and experts from the original discipline have to interpret our output coordinates. Problems arise at these two interfaces (coordination and interpretation), since the experts interested in knowledge gain give the algorithmic transformations out of their hands and thus no longer control the process of making the coordination decisions visible.

If we want to combine the disciplines, we have to find a way to incorporate expert methods (and not ``just'' research results already given) from other disciplines into our algorithms. Mathematics then provides answers on how these methods can most effectively be arranged to achieve a particular result. Like in Sec.~\ref{sec:algo}: $(a\odot c)\oplus b \oplus (b\odot c)$ needs $4$ comparisons, but it is identical to $((a\oplus b)\odot c)\oplus b$ which only needs $3$ comparisons. Arranging the terms of an expression is mathematics, computing the result of the expression is due to other disciplines. This also accounts for Sec.~\ref{sec:afs}.

The interdependence of the disciplines doesn't have to be that deep. Numerics can also take on partial tasks to a varying extent, as described in Sec.~\ref{sec:ca} and Sec.~\ref{sec:fs}. The fundamental difficulty of overemphasizing semantic dimensions of numerical values at the aforementioned interfaces remains in these cases, however. 

\paragraph{Acknowledgement.} This article has partially been financed by the Cluster of Excellence MATH+, project EF5-4 ``The Evolution of Ancient Egyptian – Quantitative and Non-Quantitative Mathematical Linguistics''.

%
%
%
\bibliographystyle{splncs04}
\bibliography{main}
\end{document}